\documentclass{article}

     \PassOptionsToPackage{numbers, compress}{natbib}
     \usepackage[final]{neurips_2021}

\usepackage[numbers]{natbib}
\newcommand{\RNum}[1]{\uppercase\expandafter{\romannumeral #1\relax}}
\usepackage[edges]{forest}
\usepackage[utf8]{inputenc} % allow utf-8 input
\usepackage[T1]{fontenc}    % use 8-bit T1 fonts
\usepackage{hyperref}       % hyperlinks
\usepackage{url}            % simple URL typesetting
\usepackage{booktabs}       % professional-quality tables
\usepackage{amsfonts}       % blackboard math symbols
\usepackage{nicefrac}       % compact symbols for 1/2, etc.
\usepackage{microtype}      % microtypography
\usepackage{amsmath,amsthm,amssymb,amsfonts}
\usepackage{amsmath,graphicx}
\usepackage{multicol}
 \usepackage{xcolor}
\usepackage{xpatch}
\usepackage{hyperref}
\usepackage{comment}
   \usepackage{xcolor}
    \usepackage{lipsum}
    \usepackage{amsmath}
\usepackage{algorithm}
\usepackage{amsthm}
\makeatother

\usepackage{hyperref}
\DeclareMathOperator*{\argminA}{arg\,min}

\AfterEndEnvironment{theorem}{\noindent\ignorespaces}

\AfterEndEnvironment{lemma}{\noindent\ignorespaces}

\def\N{\mathcal{N}}

\def\R{\mathbb{R}}

\usepackage{multirow}
\usepackage[noend]{algpseudocode}
\usepackage{amsthm}
\usepackage{enumitem}
\def\proj{\mathop{\rm proj}\nolimits}

\def\descent{\mathop{\rm descent}\nolimits}
\def\svdinit{\mathop{\rm svd-initialization}\nolimits}
\usepackage{caption} 

\usepackage{caption,subcaption}% http://ctan.org/pkg/{caption,subcaption}
 \usepackage{tikz}
\usetikzlibrary{arrows.meta,automata,positioning}
    \usepackage{mathtools}
\usepackage{amsmath,bm}
\usepackage{algorithm}
\usepackage{xcolor}
\usepackage[noend]{algpseudocode}
\def\proj{\mathop{\rm proj}\nolimits}

\def\trace{\mathop{\rm trace}\nolimits}
\def\descent{\mathop{\rm descent}\nolimits}

\usepackage{bbm}
\def\BState{\State\hskip-\ALG@thistlm}
\makeatother
\usepackage{bbm}
\usepackage{xpatch}

\makeatletter 

\NewDocumentCommand{\set}{o m}{%
  \IfNoValueTF{#1}
    {\{#2\}}
    {\{#1 \mid #2\}}%
}

\title{Learning Robust Hierarchical Patterns of Human Brain across Many fMRI Studies}

\author{%
  Dushyant Sahoo \\
  Department of Electrical Engineering\\
  University of Pennsylvania\\
  \texttt{sadu@seas.upenn.edu} \\
   \And
   Christos Davatzikos \\
   Department of Radiology \\
   University of Pennsylvania\\
   \texttt{christos.davatzikos@uphs.upenn.edu } \\
}

\begin{document}

\maketitle
\iffalse
\textbf{Todo} add experiments when the site specific data is generated from a non-linear model or some different model. Not needed now

        \textbf{Todo:} Visualization of $C_1$ and $D_1$ and show that it is different from covariance matrix i.e. it not does capture all the diagonal terms. Not needed now 

\textbf{Todo}: Check split sample reproducibility of $\textbf{D}$, do we actually have same $\mathbf{D}$? It seems that in the split sample reproducibility experiment, $\mathbf{D}$ appears to be reproducible. not needed now
\fi

\begin{abstract}
Multi-site fMRI studies face the challenge that the pooling introduces systematic non-biological site-specific variance due to hardware, software, and environment. In this paper, we propose to reduce site-specific variance in the estimation of hierarchical Sparsity Connectivity Patterns (hSCPs) in fMRI data via a simple yet effective matrix factorization while preserving biologically relevant variations. Our method leverages unsupervised adversarial learning to improve the reproducibility of the components. Experiments on simulated datasets display that the proposed method can estimate components with higher accuracy and reproducibility, while preserving age-related variation on a multi-center clinical data set. 
%[... Add some results and p-values ...]
%Experiments on simulated and real datasets display the effectiveness of the proposed formulation. 
\end{abstract}

\section{Introduction}
Multi-site fMRI studies have gained a lot of interest over the last decade \cite{noble2017multisite,di2014autism}. One reason for this is the necessity to evaluate a hypothesis in multiple settings/sites and make the hypothesis result generalizable to a diverse population. Also, the pooling of data is essential when studying rare disorders or neurological conditions where the aim is to generalize the results to diverse populations \cite{dansereau2017statistical,keshavan2016power}. However, the data pooling often results in the introduction of non-biological systematic variance due to differences in scanner hardware and imaging acquisition parameters \cite{shinohara2017volumetric}. This additional variability can lead to spurious results and a decrease in statistical power. The variability can also result in hindrance in the estimation of true biological changes or in inferring non-biological differences as biological because of the correlation between site effects and biological predictors.  Many studies working with multi-site data fMRI have reported considerable variability due site or scanner effects \cite{abraham2017deriving,jovicich2016longitudinal,noble2017multisite}. 

The non-biological variability introduced due to inter-scanner and inter-protocol variations can affect the estimation of the common features derived from fMRI \cite{yu2018statistical}, such as functional connectivity \cite{shinohara2017volumetric} or sparse hierarchical factors (this paper). These features were used for the study of the brain's function during aging \cite{raichle2015brain}, of various neurological disorders \cite{fornito2016fundamentals,stam2014modern}, and tasks \cite{cook2007aging}. The non-biological variability can considerably reduce these features' reproducibility across different datasets and hence their utility as biomarkers for diseases that disrupt functional connectivity. Thus the removal of non-biological variance introduced by pooling of the data is essential for many neuroimaging studies.

Independent Component Analysis \cite{smith2009correspondence}, Non Negative Matrix Factorization \cite{potluru2008group}, Sparse Dictionary Learning \cite{lee2010data} and Sparse factor Modeling \cite{sahoo2019sparse,sahoo2018gpu} are some of the commonly used methods for estimating interpretable components that capture the complexity of the human brain function at rest. However, these methods do not account for the hierarchical organization of the human brain \cite{doucet2011brain, demirtacs2019hierarchical}. Recently, deep learning based methods \cite{zhang2020hierarchical,huang2017modeling} have been developed to estimate hierarchical networks. However, they fail to capture overlapping components and the existence of positively and negatively correlated nodes in a component that are shown to capture the underlying structure better \cite{andersen2018bayesian,eavani2015identifying}. Hierarchical Sparse Connectivity Patterns (hSCPs) overcome these limitations by capturing interpretable hierarchical, sparse, and overlapping components using a linear deep matrix factorization approach. The method not only captures shared patterns but also the subject-specific weights of these patterns, thus capturing the heterogeneity of brain activity patterns across individuals. In this paper, we focus on robust estimation of hSCPs \cite{eavani2015identifying,sahoo2020hierarchical,sahoo2021extraction} in a multi-site regime. 

Many existing methods to reduce site effects are based on an empirical Bayes method ComBat \cite{johnson2007adjusting}, which was developed to remove `batch effects' in genetics and has been applied for harmonizing different measures derived from structural \cite{pomponio2020harmonization, fortin2017harmonization} and functional MRI \cite{yu2018statistical}. But these harmonization methods can not be directly applied to the hSCP method because of the loss in the structure of captured heterogeneity (more details in section \ref{sec:hSCP}). See Appendix \ref{sec:rel_work} for extended discussion on related work.
%Failure to reduce the site variability can lead to incorrect conclusions when the components are used for association with demographic variables, clinical phenotypes, neurological diseases, etc. Thus it is desirable to develop methods that can reduced site effects and improve the reproducibility of the hierarchical components. 

\paragraph{Contribution.} In this paper, we develop a new model that is robust to site-effects in the estimation of sparse hierarchical connectivity pattern components (rshSCP). For this, the method learns site-specific features and global space, storing the information about the scanner and site, and uses these features to reduced site effects in the components. We also use an existing adversarial learning approach \cite{sahoo2021extraction}  on top of our method to improve the reproducibility and generalizability of the components across components from the same site. We formulate the method as a non-convex optimization problem which is solved using stochastic gradient descent. Experiments on simulated and real datasets show that our method can improve the split-sample and leave one site reproducibility of the components while retaining age-related biological variability in the data, thus capturing informative heterogeneity.

\section{Method}
%\subsection{... Subsection title missing ...}
% just an introduction to the method section, subsection title is not necessary here
%\section{Overview}
%We aim at extracting hierarchical Sparse Connectivity Patterns (hSCP) derived from 3D resting-state functional magnetic resonance imaging (fMRI) time series acquired at multiple sites.
%Functional connectivity is captured as an undirected graph in which the nodes represent functional networks and the edges are the correlation between the average signal within these networks.
%The functional connectivity is represented as a symmetric positive definite matrix.
%We aim at decomposing the connectivity matrix into reproducible hierarchical components that are largely invariant to site while preserving biological information.

%To obtain these, we extend an adversarial formulation of hSCP with site-specific component.

In this section, we first present a brief overview of the hierarchical Sparse Connectivity Pattern (hSCP) and its adversarial formulation. We then describe our main method and the joint formulation incorporating adversarial learning. We follow the notation in \cite{sahoo2020hierarchical} and \cite{sahoo2021extraction}. The set of symmetric positive definite matrices of size $P \times P$ is denoted by $\mathbb{S}^{P \times P}_{++}$. Matrix $\mathbf{A}$ with all the elements greater than or equal to $0$ is denoted by $\mathbf{A} \geq 0$. $\mathbf{J}_P$ denotes $P \times P$ matrix with all elements equal to one. $P \times P$ identity matrix is denoted by $\mathbf{I}_P$ and element-wise product between two matrices $\mathbf{A}$ and $\mathbf{B}$ is denoted by $\mathbf{A} \circ \mathbf{B}$.
%$\mathbf{A} \geq 0$ denotes all the elements of the matrix $\mathbf{A}$ are greater than or equal to $0$. $\mathbf{J}_P$ denotes $P \times P$ matrix with all elements equal to one. $\mathbf{I}_P$ denotes $P \times P$ identity matrix. $| \mathcal{I}|$ denotes cardinality of set $\mathcal{I}$. $\mathbf{A} \circ \mathbf{B}$ is the element wise product between two matrices $\mathbf{A}$ and $\mathbf{B}$. 
\subsection{Introduction to hierarchical Sparse Connectivity Patterns}
\label{sec:hSCP}
Hierarchical Sparse Connectivity Patterns (hSCP) \cite{sahoo2020hierarchical} is a hierarchical extension of Sparse Connectivity Patterns (SCPs), first defined by \cite{eavani2015identifying} to estimate sparse functional patterns in the human brain using fMRI data. Let there be $N$ number of subjects or participants, and each subject's BOLD fMRI time series has $T$ time points and $P$ nodes representing regions of interest. The input to hSCP are correlation matrices $\mathbf{\Theta}^n \in \mathbb{S}^{P \times P}_{++}$ where $i$th and $j$th element of the matrix is the correlation between time series of $i$th and $j$th node. hSCP then outputs a set of shared hierarchical patterns following the below equations:
 \begin{align*}
\mathbf{\Theta}^n \approx \mathbf{W}_1\mathbf{\Lambda}_1^n \mathbf{W}_1^\top, \quad \ldots \quad \mathbf{\Theta}^n \approx \mathbf{W}_1\mathbf{W}_2 \ldots\mathbf{W}_K \mathbf{\Lambda}^n_K \mathbf{W}_K^\top\mathbf{W}_{K-1}^\top \ldots \mathbf{W}_1^\top ,
 \end{align*}
 where $\mathbf{\Lambda}^n_k$ is a diagonal matrix having positive elements storing relative contribution of the components for the $n$th subject at $k$th level, $K$ is the depth of hierarchy and $P > k_1 > \ldots > k_K $ i.e. each successive level in the hierarchy has less number of components than the previous one. In the above formulation, $\mathbf{W}_1 \in \mathbb{R}^{P \times k_1}$ stores $k_1$ components at the bottom most level, and each successive multiplication by $\mathbf{W}_2$, $\mathbf{W}_3$, $\ldots$, $\mathbf{W}_K$ linearly transforms to a lower dimensional space of $k_2$, $k_3$, $\ldots$, $k_K$ dimension. Let $\mathcal{W} = {\set[\mathbf{W}_r]{r=1,\ldots, K}}$ be the set storing sparse components shared across all subjects and $\mathcal{D}= {\set[\mathbf{\Lambda}_r^n]{r=1,\ldots, K; n=1\ldots,N}}$ be set storing subject specific diagonal matrix with $\mathbf{\Lambda}^n_r \geq 0$.  The hierarchical components are estimated by solving the below optimization problem:
  \begin{equation}
\begin{aligned}
& \underset{\mathcal{W},\mathcal{D}}{\text{min}}
& & H(\mathcal{W},\mathcal{D},\mathcal{C}) =  \sum_{n=1}^{N} \sum_{r=1}^{K}\|\mathbf{\Theta}^n - (\prod_{j=1}^{r}\mathbf{W}_j)\mathbf{\Lambda}_r^n(\prod_{j=1}^{r}\mathbf{W}_j)^\top \|_F^2\\
& \text{s.t.}
&&\|\mathbf{w}^r_l\|_1 < \tau_r,\; \|\mathbf{w}^r_l\|_\infty \leq 1, \; \trace(\mathbf{\Lambda}_r^n) =1,  \\
&&&\mathbf{\Lambda}_r^n \geq 0, \; \forall n,r,l; \hspace{2em} \mathbf{W}_j \geq 0, \; j=2,\ldots,K,
\end{aligned}
   \label{problem:hscp}
 \end{equation}
where $\mathcal{C} = {\set[\mathbf{\Theta}^n]{n=1,\ldots, N}}$, $l=1,\ldots,k_r, \; n=1,\ldots,N \; \text{and} \; r=1,\ldots,K$. $L_1$, $L_{\infty}$ and $\trace$ constraints help the problem to identify a decomposition which can provide reproducible components. More details can be found in the original paper \cite{sahoo2020hierarchical}. We will be denoting above constraint set as $ \Omega_{\mathcal{W}} = \set[\mathbf{W}]{\|\mathbf{w}^r_l\|_1 < \tau_r,\; \|\mathbf{w}^r_l\|_\infty \leq 1,  \mathbf{W}_j \geq 0, \; j=2,\ldots,K}  $ and $\Psi = \set[\mathbf{\Lambda}]{\trace(\mathbf{\Lambda}_r^n) =1,\mathbf{\Lambda}_r^n \geq 0}$. A detailed description of the method can be found in \cite{sahoo2020hierarchical}.

\paragraph{Can we use standard harmonization approaches?} These methods reduce site effects by adjusting for additive and multiplicative effects for each feature in data separately and use emperical Bayes estimates the model parameters. These methods can be used in the case of hSCP in two ways. First, harmonization can be directly applied to each element of the correlation matrices, which is the input of hSCP. This will reduce site effects from each element of the correlation matrix, thus from the complete input, but the final matrix that does not necessarily follow the essential property of a correlation matrix i.e., positive definiteness. For similar reasons, COMBAT can not be directly applied to time series; if applied, it can change the inference derived from the correlation matrix. Second, harmonization can be directly applied to $\mathbf{\Lambda}$ to remove site effects. To understand this, we look at the hSCP formulation at one level:
\begin{align*}
  \mathbf{\Theta}^n \approx \sum_{l=1}^{k}d^n_l \mathbf{w}_l \mathbf{w}_l^\top   \approx \mathbf{W}\mathbf{\Lambda}^n\mathbf{W}^\top,
\end{align*}
where $d^n_l$ are non-zero elements storing the subject-specific information, which can be affected by the variability introduced by the site. In this model, harmonizing each feature across different sites will change the relative contribution of the components in each subject's functional structure, which will remove the interpretability of the subject-wise weights, which is not desirable. Instead, a two step optimization procedure can be used to incorporate ComBat with hSCP (ComBat hSCP). We first run hSCP and use ComBat on the extracted $\mathbf{\Lambda}^n$  to get harmonized subject specific information $\mathbf{\Delta}^n \in \mathbb{R}^{k_1 \times k_1}$ for each subject. We then re-fitted 
$\mathbf{W}$ using the below decomposition-
\begin{align*}
  \mathbf{\Theta}^n \approx  \mathbf{W}(\mathbf{\Delta}^n + \mathbf{S})\mathbf{W}^\top.
\end{align*}
We added a diagonal shift matrix $\mathbf{S} \in \mathbb{R}^{k_1 \times k_1}$ such that $\mathbf{\Delta}^n + \mathbf{S}$ is positive for each subject and performed the optimization to estimate $\mathbf{W}$ and $\mathbf{S}$. We show through experiments that this baseline two step optimization procedure is not optimal and performs worse than vanilla hSCP. 
\subsection{Adversarial Learning in hSCP}
\label{sec:adv_hscp}
Adversarial learning has shown to achieve state-of-the-art performance of various matrix factorization approaches \cite{he2018adversarial,luo2020adversarial}. Recently, \citet{sahoo2021extraction} demonstrated that incorporating adversarial learning in the estimation problem of hSCP can improve the reproducibility of the hierarchical components. The method is based on perturbation of input data $\mathbf{\Theta}^n$ to learn stable components robust to adversaries. First the input data is perturbed to generate new data $\mathbf{\Gamma}^n = \mathbf{\Theta}^n + \sigma \mathbf{J}_P$ where $\sigma$ is the standard deviation of the data. We consider this perturbation as rank one perturbation which will transform eigenvalues of each subject’s correlation matrix differently. Since the addition of ones does not change the symmetric property of the matrix, we just scale the matrix such that the diagonal matrix contains $1$ and by Weyl's inequality about perturbation \cite{stewart1998perturbation} it will be positive definite. We have experimented with randomly positively scaled rank one perturbation, but the results were worse than the vanilla hSCP. Also, using rank one perturbation, it is easier to control eigenvalues (and noise added) of the modified matrix than using rank k perturbation. The perturbed set of components $\mathbf{\tilde{W}}_1$ are then estimated using the new data and are ensured to be close to $\mathbf{{W}}_1$ by solving the below minimization problem:
    \begin{align}
       A(\mathbf{\hat{W}}_1) = \alpha\| \mathbf{\tilde{W}}_1 - \mathbf{W}_1\|_F^2 +  \sum_{n=1}^{N} \|\mathbf{\Gamma}^n  - \mathbf{\tilde{W}}_1 \mathbf{\Lambda}^n \mathbf{\tilde{W}}^\top_1 \|_F^2.
       \label{eq:attack}
    \end{align}
In the above optimization problem, the first term constraints $\mathbf{\tilde{W}}_1$ to be close to $\mathbf{{W}}_1$ and second term is used for learning the components using the perturbed data. Aim of the learner is to estimate $\mathbf{\Lambda}^n$ and $\mathbf{W}_1$ by minimizing the below cost function:
    \begin{align}
     D(\mathbf{W},\mathbf{\Lambda}) =    \sum_{n=1}^{N} \|\mathbf{\Theta}^n  - \mathbf{\tilde{W}}_1\mathbf{\Lambda}^n \mathbf{\tilde{W}}^\top_1 \|_F^2 +\beta \sum_{n=1}^{N}\|\mathbf{\Theta}^n  - \mathbf{W}_1\mathbf{\Lambda}^n \mathbf{{W}}^\top_1 \|_F^2 ,
     \label{eq:defense}
    \end{align}
    for a fixed $\mathbf{\tilde{W}}_1$. Learner first estimates subject specific information $\mathbf{\Lambda}^n$ using perturbed weight matrix and then use it to learn $\mathbf{{W}}_1$. Below is the complete optimization problem for adversarial learning of hSCP:
  \begin{equation}
\begin{aligned}
& \underset{\mathcal{W},\mathcal{D}}{\text{min}}
& &  H(\mathcal{\tilde{W}},\mathcal{D},\mathcal{C}) + \beta H(\mathcal{W},\mathcal{D},\mathcal{C}) \\
& \text{s.t.}
&& \mathbf{\tilde{W}}_r = \argminA_{\mathbf{\hat{W}}_r} \alpha\| \mathbf{\hat{W}}_r - \mathbf{{W}}_r\|_F^2 + H(\mathcal{\tilde{W}},\mathcal{D},\mathcal{P}) \quad r=1,\ldots,K \\
&&& \mathbf{\tilde{W}}_r, \mathbf{{W}}_r \in \Omega, \qquad \mathcal{D} \in \Psi,
\end{aligned}
   \label{problem:h.adv}
 \end{equation}
where $\mathcal{\tilde{W}} = {\set[\mathbf{\tilde{W}}_r]{r=1,\ldots, K}}$ and $\mathcal{P} = {\set[\mathbf{\Gamma}^n]{n=1,\ldots, N}}$. We next discuss our main method which aims to reduce the site effects.
 
 \subsection{Robust to site hSCP}
Estimating hSCPs in multi-site data can introduce non-biological variances in the components and the subject-specific information. One of the typical approaches would be to use harmonization methods mentioned previously, but it would lead to a loss in the structure of these features, which in turn will lose interpretability. 
Instead of removing site effects after estimating the components, we jointly model the sparse components and the site effects, and estimate robust to site hSCP (rshSCP).
We first look at the case when there is only one level in the hierarchy, which then can be extended to multiple levels.
Let there be total $S$ sites, $\mathcal{I}_s$ be the set storing subjects from site $s$ and $\mathbf{y} \in \mathbb{R}^{N \times S}$ be the one-hot encoded site labels
We hypothesize that there is a space $\mathbf{V} \in \mathbb{R}^{P \times P}$ storing site and scanner information for all the possible available data, and for each site $s$, we have space $\mathbf{U}^s \in \mathbb{R}^{P \times P}$ storing site-specific information for $s=1,\ldots, S$.
Based on the above hypothesis, we decompose the correlation matrix $\mathbf{\Theta}^{n}$ of $n \in  \mathcal{I}_s$ to jointly estimate the hSCPs, $\mathbf{U}^s$ and $\mathbf{V}$ as:
\begin{align}
    \mathbf{\Theta}^{n} \approx \underbrace{\mathbf{W} \mathbf{\Lambda}^{n} \mathbf{W}^\top}_{\substack{\text{decomposition of} \\ \text{subject components}}} + \underbrace{\mathbf{U}^s\mathbf{V}}_{\substack{\text{decomposition of} \\ \text{site components}}}, 
    \label{eq:site}
\end{align}
where $\mathbf{U}^s$ is constrained to be a diagonal matrix and $ L_1$ sparsity constraint is used for $\mathbf{V}$ to prevent overfitting. In addition to estimating the site effects, we train the model such that the predictive power of subject-specific information $\mathcal{D}$ for predicting site is reduced, which can assist in removing site information. For this, we train a differentiable classification model $F(\zeta, \mathcal{D})$ parameterized by $\zeta$ with input $\mathbf{\Lambda}^{n}$ that return site predictions $\hat{\mathbf{y}} \in \mathbb{R}^{N \times S}$. These predictions indicate the probabilities that each of $N$ inputs belongs to each of $S$ site labels. The classification model is trained by optimizing for $\zeta$ such that the cross-entropy loss $\mathcal{L}(\zeta, \mathcal{D},  \mathbf{y})$ between the predictions $\hat{\mathbf{y}}$ and the true site labels $\mathbf{y}$ is minimized:
\begin{align}
 \zeta^* =  \underset{\zeta}{\text{arg min}}  -\frac{1}{N}\sum_{n=1}^N \sum_{s=1}^{S}y_{n,s}\log \hat{y}_{n,s}.
    \label{eq:1}
\end{align}
 Using this model, we modify $\mathbf{\Lambda}^{n}$ such that its predictability power reduces. We achieve this by maximizing the above loss with respect to $\mathcal{D}$. This will result in a minimax game, where the classifier learns to minimize the cross-entropy or the surrogate classification loss, and $\mathcal{D}$ is adjusted to maximize the loss. The joint optimization problem can be written as:
 \begin{equation}
\begin{aligned}
\underset{\zeta}{\text{max}} \; &\underset{\mathbf{W},\mathcal{D},\mathcal{U},\mathbf{V}}{\text{min}} \quad && \sum_{s=1}^S \sum_{n \in \mathcal{I}_s}   \|\mathbf{\Theta}^{n} - \mathbf{W} \mathbf{\Lambda}^{n} \mathbf{W}^\top - \mathbf{U}^s\mathbf{V}\|_F^2 - \gamma \mathcal{L}(\zeta, \mathcal{D}, \mathbf{y}) \\
& \hspace{1em} s.t. && \mathbf{W} \in \Omega, \quad \mathcal{D} \in \Psi, \quad \|\mathbf{v}_p\|_1 < \mu, \; p=1,\ldots,P,
\end{aligned}
\label{eq:site_ind}
\end{equation}
where $\mathcal{U} = \{U_s|s=1,\ldots,S \}$, $\mathbf{v}_p$ is the $p$th column of $\mathbf{V}$.
\subsection{Complete Model}
We can combine the above formulation (\ref{eq:site_ind}) at multi level with the adversarial learning (\ref{problem:h.adv}) to jointly model hSCPS and site effects. Let 
\begin{align}
G(\mathcal{W},\mathcal{D},\mathcal{C}) =   \sum_{s=1}^S \sum_{n \in \mathcal{I}_s} \sum_{r=1}^{K}\|\mathbf{\Theta}^n - (\prod_{j=1}^{r}\mathbf{W}_j)\mathbf{\Lambda}_r^n(\prod_{n=1}^{r}\mathbf{W}_n)^\top - \mathbf{U}^s_r \mathbf{V}_r \|_F^2,     
\end{align}
then the joint optimization problem can be written as:
\begin{equation}
\begin{aligned}
\underset{\zeta}{\text{max}} \;
& \underset{\mathcal{W},\mathcal{D},\mathcal{U},\mathbf{V}}{\text{min}}
& & J(\mathcal{\tilde{W}},\mathcal{W},\mathcal{D},\mathcal{C})= G(\mathcal{\tilde{W}},\mathcal{D},\mathcal{C}) + \beta G(\mathcal{W},\mathcal{D},\mathcal{C}) + \gamma \mathcal{L}(\zeta, \mathcal{D}, \mathbf{y}) \\
& \text{s.t.}
&& \mathbf{\tilde{W}}_r = \argminA_{\mathbf{\hat{W}}_r} \alpha\| \mathbf{\hat{W}}_r - \mathbf{{W}}_r\|_F^2 + G(\mathcal{\tilde{W}},\mathcal{D},\mathcal{P}) \quad r=1,\ldots,K \\
&&& \mathbf{\tilde{W}}_r,\mathbf{{W}}_r \in \Omega \quad \mathcal{D} \in \Psi, \quad \|\mathbf{v}_p\|_1 < \mu, \; p=1,\ldots,P,
\end{aligned}
   \label{problem:complete}
 \end{equation}

The optimization problem defined above is a non-convex problem that we solve using alternating minimization. Complete algorithm and the details about the optimization are described in Appendix \ref{sec:alg}. Note that the random initialization of the variables can result in a very different final solution that might be far from the ground truth. One such solution for $\mathbf{U}$ and $\mathbf{V}$ would be the identity matrix since all the correlation matrices have one as their diagonal element, which can drastically change the final components. It might also be possible that $\mathbf{V}$ might store highly reproducible components since they are present in most individuals, leading to a decrease in reproducibility of hSCPs.  We prevent these cases by using $\svdinit$ \cite[Algorithm 2]{sahoo2020hierarchical} for $\mathcal{W}$ and $\mathcal{D}$, where, in the starting, most of the variability associated with data is stored in $\mathcal{W}$ and $\mathcal{D}$. In this way, we can prevent $\mathbf{V}$ from storing highly reproducible components during initial iterations. We initialize $\mathbf{U}^s$ and $\mathbf{V}$ using the below equation:
\begin{align}
    \mathbf{U}^s_r = \left[ \frac{1}{|\mathcal{I}_s|}\left( \sum_{n \in \mathcal{I}_s} \mathbf{\Theta}^n - (\prod_{j=1}^{r}\mathbf{W}_j)\mathbf{\Lambda}_r^n(\prod_{n=1}^{r}\mathbf{W}_n)^\top \right)\mathbf{J}_{p }\right] \circ \mathbf{I}_p, \qquad \mathbf{V}_r =  \frac{1}{P}\mathbf{J}_{P}.
    \label{eq:init}
\end{align}
This complete initialization procedure ensures that the algorithm starts with the majority of variability in the data stored in $\mathcal{W}$ and $\mathcal{D}$, and $\mathbf{U}^s$ start from the residual variance left in site $s$ after the $\svdinit$ procedure. We show in the next sections that this simple strategy, though sub-optimal, can help estimate reproducible components with diminished site effects. All the code is implemented in \textsc{MATLAB} and will be released upon publication.

\section{Experiment}
\subsection{Simulated Dataset}
\label{sec:simu}

\paragraph{One level.}We first generate simulated dataset at one level to evaluate the performance of our model against the vanilla hSCP. We simulate data with $p=50$, $k_1=10$, $S = 4$ with $200$, $300$, $400$ and $500$ number of participants in each site. We generated sparse shared components $\mathbf{W}_1$ with percentage of non-zeros equal to $60\%$ and each element sampled from $\N(0,1)$. We then generate correlation matrix for $n$th subject belonging to $s$th site using:
\begin{align}
   {\mathbf{\Theta}}^{n} = \left(\mathbf{W}_{1}+\mathbf{E}_1^n \right)\mathbf{\Lambda}^{n} \left(\mathbf{W}_{1}+\mathbf{E}_2^n\right)^\top + \mathbf{U}^s\mathbf{V} + \mathbf{E}_2^{n},
\end{align}
where $\mathbf{U}^s$ is a diagonal matrix with positive elements sampled from $\N(1,.1)$, $\mathbf{V}$ is a random matrix sampled from wishart distribution, each element of $\mathbf{\Lambda}^{n} $ is sampled from $\N(4,1)$ and $\mathbf{E}_1^{n}$ is the noise matrix added to the components whose each element is sampled from $\N(0,.1)$ and $\mathbf{E}_2^{n}$ is added to ensure that the final matrix is positive definite. However the diagonal elements of ${\mathbf{\Theta}}^{n}$ are not equal to $1$. To make them $1$, we extract diagonal elements $\mathbf{D}$ of ${\mathbf{\Theta}}^{n}$ and get the new correlation matrix as $\mathbf{D}^{1/2}{\mathbf{\Theta}}^{n}\mathbf{D}^{1/2}$. We used a feed-forward neural network for the classification model with two hidden layers. The networks contain the following layers: a fully connected layer with $50$ hidden unites, dropout layer with rate $0.2$, ReLU, a fully-connected layer with $4$ hidden units and a softmax layer. Optimal value of hyperparameters $\alpha$, $\beta$, $\mu$ and $\tau_1$ are selected from $[0.1,1]$, $[1,5]$, $[0.1,0.5,1]$ and $10^{[-2:2]}$. The criterion for choosing the best hyperparameter is maximum split-sample reproducibility. The split sample reproducibility is the normalized inner product between the components estimated on two random equal splits of the data. Split sample reproducibility tries to answer the question of whether the components are generalizable across subjects from the same sites or not. We compared different methods for estimation of hierarhical components- hSCP, ComBat hSCP, hSCP with adversarial learning (Adv. hSCP), rshSCP, rshSCP with adversarial learning (Adv. rshSCP), rshSCP and Adv. rshSCP with random initialization (rshSCP w/ rand. and Adv. rshSCP w/ rand.). Table \ref{tbl:simu_repro} shows the reproducibility of the components generated from different methods. It is computed over $15$ runs in all the experiments. We used accuracy of the estimated components as a performance measure. It is defined as the normalized inner product between ground truth components and estimated components. All the experiments were run on a four i7-6700HQ CPU cores single ubuntu machine.
\begin{table}[t!]
\caption{Accuracy of the components on simulated dataset at one level. \label{tbl:simu_accu}}
\centering
        \begin{tabular}{ lcccc }
 %\hline
 %\multicolumn{5}{|c|}{Comparison by varying number of components}\\
 \toprule
 {Method} & {$k_1=8$} & {$k_1=10$}  & {$k_1=12$} & {$k_1=14$} \\
 \midrule
hSCP & $0.789$    & $0.787$ &   $0.745$ & $0.736$\\
ComBat hSCP & $0.763$    & $0.759$ &   $0.731$ & $0.718$\\
Adv. hSCP & $0.869 $     & $0.875$ &  $0.862$  & $0.854$ \\
rshSCP & $0.873 $     & $0.865$ &  $0.843$  & $0.867$ \\
Adv. rshCP & $\mathbf{0.903} $ &$ \mathbf{0.910}$  &  $\mathbf{0.902}$  &  $\mathbf{0.908}$\\
rshSCP w/ rand. & $ 0.856\pm0.039 $     &$0.834\pm0.055$  &  $0.824 \pm 0.031$  &  $0.818\pm0.036$ \\
Adv. rshSCP. w/ rand. & $ 0.897\pm0.030 $     &$0.895\pm0.036$  &  $0.892 \pm 0.023$  &  $0.886\pm0.034$\\
 \bottomrule
\end{tabular}
\end{table}
\begin{table}[t!]
\caption{Split sample reproducbility on simulated dataset at one level. \label{tbl:simu_repro}}
\centering
        \begin{tabular}{ p{2.97cm}cccc }
 %\hline
 %\multicolumn{5}{|c|}{Comparison by varying number of components}\\
 \toprule
 {Method} & {$k_1=8$} & {$k_1=10   $}  & {$k_1=12$} & {$k_1=14$} \\
 \midrule
hSCP & $0.769\pm0.052$    & $0.798 \pm 0.047$ &   $0.739 \pm 0.053$ & $0.734 \pm 0.047$\\
ComBat hSCP & $0.749\pm0.040$    & $0.750 \pm 0.052$ &   $0.724 \pm 0.049$ & $0.719 \pm 0.052$\\
Adv. hSCP  & $0.781 \pm 0.037$     & $0.818 \pm 0.031$ &  $0.780 \pm 0.034$  & $0.750 \pm 0.031 $ \\
rshSCP & $0.825 \pm 0.039$     & $0.845\pm 0.030$ &  $\mathbf{0.826\pm 0.039}$  & $0.779\pm 0.036$ \\
Adv. rshSCP & $ \mathbf{0.840 \pm 0.044}$     &$\mathbf{0.869 \pm 0.034}$  &  $0.815 \pm 0.035$  &  $\mathbf{0.802 \pm 0.030}$\\
rshSCP w/ rand. & $0.804 \pm 0.085$     & $0.818\pm 0.086$ &  $0.780\pm 0.068$  & $0.758\pm 0.071$ \\
Adv. rshSCP w/ rand. & $ 0.826 \pm 0.069$     &$0.833 \pm 0.081$  &  $0.801 \pm 0.074$  &  $0.782 \pm 0.077$\\
 \bottomrule
\end{tabular}
\end{table}

\begin{table}[t!]
\caption{Accuracy of the components on hierarchical simulated dataset.  \label{tbl:simu_accu_new}}
\centering
        \begin{tabular}{ lcccccccc }
        &  \multicolumn{4}{c}{$k_2 = 4$}     & \multicolumn{4}{c}{$k_2 = 6$}              \\
    \cmidrule(r){2-5} \cmidrule(r){6-9}
 %\hline
 %\multicolumn{5}{|c|}{Comparison by varying number of components}\\
 {Method \textbackslash \hspace{0.2em} $k_1$} & {$8$} & {$10$}  & {$12$} & {$14$} & {$8$} & {$10$}  & {$12$} & {$14$}\\
 \midrule
hSCP & $0.806$    & $0.801$ &   $0.783$ & $0.777$  & $0.797$    & $0.790$ &   $0.773$ & $0.766$ \\
ComBat hSCP & $0.788$    & $0.776$ &   $0.743$ & $0.729$  & $0.779$    & $0.766$ &   $0.747$ & $0.734$ \\
Adv. hSCP & $0.875 $     & $0.872$ &  $0.870$  & $0.864$ & $0.863$    & $0.859$ &   $0.849$ & $0.851$ \\
rshSCP & $0.881 $     & $0.876$ &  $0.860$  & $0.862$  & $0.874$    & $0.871$ &   $0.852$ & $0.859$ \\
Adv. rshSCP & $ \mathbf{0.904} $     &$ \mathbf{0.909}$  &  $\mathbf{0.904}$  &  $\mathbf{0.907}$  & $\mathbf{0.902}$    & $\mathbf{0.903}$ &   $\mathbf{0.904}$ & $\mathbf{0.902}$\\
 \bottomrule
\end{tabular}
\end{table}
\paragraph{Accuracy.} Table \ref{tbl:simu_accu} displays the accuracy of different methods on the simulated dataset. Here, accuracy is defined as the average correlation between estimated components and the ground truth components. From the results, we can see that the rshSCP with adversarial learning can significantly improve the components' accuracy and the reproducibility of the components. The baseline (ComBat hSCP) performs worse than vanilla hSCP. One reason for this might be that the harmonized $\mathbf{\Lambda}$ extracted using ComBat might not necessarily result in optimal highly reproducible $\mathbf{W}$. This result bolsters our method that we need a joint optimization procedure to obtain $\mathbf{W}$ and $\mathbf{\Lambda}$ with reduced site effects. The results using random initialization instead of using the initialization strategy mentioned in the previous section indicates that random initialization brings significant variability to performance. On average, it performs worse than our strategy, but there might be instances where the random initialization can perform better, which might suggest that there might be some better strategy for initialization. Also, for $\mathbf{V}$, there is an optimal sparsity value, which achieves the best result. If $\mathbf{V}$ is dense, then it might remove essential information that might reduce reproducibility, and if it is too sparse, then we might not have desired effects to make the model robust. The results showing the variation in the accuracy and reproducibility with the sparsity of $\mathbf{V}$ are in Appendix \ref{sec:supp_simu}.
\paragraph{Site prediction.}To check if the estimated subject information ($\mathbf{\Lambda}$) has reduced predictive power to predict the site to which the subject belonged, we performed a $5$ fold cross-validation using SVM with RBF kernel. We also ran our experiment using two different feed forward networks with two different architectures: (a) a fully connected layer with  hidden units, dropout layer with the rate , ReLU, a fully-connected layer with  hidden units and a softmax layer and (b) a fully connected layer with  hidden units, dropout layer with the rate , ReLU, a fully-connected layer with  hidden units, dropout layer with rate , ReLU, a fully-connected layer with  hidden units and a softmax layer. Our model leads to a decrease in average cross-validation accuracy from $97.6\%$ to $67\%$ for SVM, $98.1\%$ to $67.3\%$ for neural network with architecture (a) and $98.2\%$ to $66.9\%$ for neural network with architecture (b). This suggests that our model can reduce the prediction capability to predict site. More details about the result are in Appendix \ref{sec:supp_simu}.

\paragraph{Two level.} Under the same settings as defined above, we generate correlation matrix from two level components with $k_2 = 4$ using:
\begin{equation}
\begin{aligned}
   {\mathbf{\Theta}}^{n} & = \tilde{\mathbf{W}}_{1}\tilde{\mathbf{W}}_{2}\mathbf{\Lambda}^{n} \tilde{\mathbf{W}}_{2}^\top \tilde{\mathbf{W}}_{1}^\top + \mathbf{U}^s\mathbf{V} + \mathbf{E}_3^{n}, \\
   \tilde{\mathbf{W}}_1 & = {\mathbf{W}}_1 \mathbf{E}_1^{n}, \quad  \tilde{\mathbf{W}}_2 = {\mathbf{W}}_2 + \mathbf{E}_2^{n}
\end{aligned}
\end{equation}
where each element of $\mathbf{W}_{2}$ is sampled from $\N(0,1)$, the percentage of non-zeros equal to $40\%$, $\mathbf{E}_1^{n}$ and $\mathbf{E}_2^{n}$ is the noise added to the components whose each element is sampled from $\N(0,.1)$ and $\mathbf{E}_3^{n}$ is added to ensure that the final matrix is positive definite. Table \ref{tbl:simu_accu_new} shows the accuracy for different values of $k_1$ and $k_2$. Selection of hyparameter is same as in the previous paragraph. We can see that the proposed method estimates most accurate ground truth components. Reproducbility results and random initialization results are availble in Appendix \ref{sec:supp_simu}.

\subsection{Real Dataset}
\label{sec:real_dataset}
\paragraph{Data.}
We collected functional MRI data from $5$ different multi-center imaging studies- 1) Baltimore Longitudinal Study of Aging (BLSA) \cite{armstrong2019predictors, resnick2003longitudinal}, the Coronary Artery Risk Development in Young Adults study
(CARDIA) \cite{friedman1988cardia}, UK BioBank (UKBB) \cite{sudlow2015uk}, Open
access series of imaging studies (OASIS) \cite{marcus2007open} and Aging Brain Cohort Study (ABC) from Penn Memory Center \cite{pluta2012vivo}. Although UK Biobank has more than $20000$ scans, we only used $2023$ randomly selected scans to avoid estimating the results that would be heavily influenced by the UK Biobank. We projected the data into a lower-dimensional space such that the number of nodes in each subject's data was $100$. Table \ref{tbl:summary} summarizes the number of participants in each site and age distribution. More details on the dataset and the preprocessing pipeline are given in Appendix \ref{sec:pre}. CARDIA data is divided into three parts because of the acquisition at three different sites.
\begin{table}[t]
 \caption{Summary characteristics of the real dataset.}
 \label{tbl:summary}
\centering
\begin{tabular}{ lcccc }
 \hline

Data Sites  & Participants & \% of Females & Age Range (Median) & Scanner \\
 \hline
 BLSA-3T    & $784$ & $56.5$ & $[22,95](68)$  & 3T
Philips \\
 CARDIA1  & $199$  & $55.7$ & $[42,61](52)$ & 3T
Siemens Tim Trio\\
 CARDIA2   & $321$  & $51.4$ & $[43,61](52)$ & 3T Philips Achieva \\
 CARDIA3   & $278$  & $55.3$ & $[43,62](52)$ & 3T Philips Achieva\\
 UKBB  & $2023$& $55.2$ & $[45,79](63)$ & 3T
Siemens Skyra\\
OASIS   & $847$ & $56.0$ & $[42.6,97](70)$&  1.5T Siemens Vision \\
ABC    & $279$ & $59.1$ & $[23,95](70)$&  3T Siemens Tim Trio\\
 \hline

\end{tabular}

\end{table}

\paragraph{Reproducibility.}
Since we don't have access to ground truth here, we compare the methods based on the split sample and leave one site reproducibility. Leave one site out reproducibility is defined as the similarity between components derived from the site $s$ and all sites except $s$. Split sample reproducibility tries to answer the question of whether the components are generalizable to other sites or not. For estimating rshSCP with only one site, we used $\mathbf{V}$ estimated from all sites except $s$ since the idea behind $\mathbf{V}$ was to store information about the site/scanner from various sites. This would also help analyze the generalization power of $\mathbf{V}$. The optimum value of the hyperparameters is selected from the range defined in section \ref{sec:simu}. $\tau_1$ and $\tau_2$ are selected from $10^{[-2:2]}$ based on maximum split-sample reproducibility. 
%fixed to be $10$ and $5$ respectively for estimating sparse components. 
The criterion for choosing the best value is the maximum split sample reproducibility. Table \ref{tbl:real_repro_h1} shows the split sample reproducibility for varied values of $k_1$ and $k_2 = 4$. Leave one site out reproducibility results are shown in Table \ref{tbl:leave_repro_h1}. Table \ref{tbl:real_repro_h2} and Table \ref{tbl:leave_repro_h2} in  \ref{sec:supp_real} shows split sample reproducibility and leave one site out reproducibility respectively at two-level for $k_2 = 6 $. The results demonstrate that the proposed method can significantly improve the split sample reproducibility and leave one site out reproducibility. For the remaining paper, we focus on the comparison between components learned using adversarial learning from hSCP and Adv. rshSCP.%
\begin{table}[t!]
\caption{Split-sample reproducbility on real dataset ($k_2 = 4$). \label{tbl:real_repro_h1}}
\centering
        \begin{tabular}{ lcccc }
 %\hline
 %\multicolumn{5}{|c|}{Comparison by varying number of components}\\
 \toprule
 {Method} & {$k_1=10$} & {$k_1=15$}  & {$k_1=20$} & {$k_1=25$} \\
 \midrule
hSCP & $0.713\pm0.039$   & $0.707\pm0.038$   &  $0.697\pm0.035$ &   $0.683\pm0.036$\\
ComBat hSCP & $0.673\pm0.049$   & $0.641\pm0.051$   &  $0.639\pm0.031$ &   $0.611\pm0.038$\\
Adv. hSCP & $0.737\pm0.041$ &  $0.719\pm0.033$  &  $0.715\pm0.037$ &    $0.710\pm0.043$ \\
rshSCP & $0.806\pm0.036$ &   $0.768\pm0.032$ &    $0.742\pm0.033$ &   $0.743\pm0.044$ \\
Adv. rshSCP & $\mathbf{0.808\pm0.030}$ & $\mathbf{0.772\pm0.036}$ & $\mathbf{0.747\pm0.034}$ &  $\mathbf{0.746\pm0.036}$ \\
 \bottomrule
\end{tabular}
\end{table}
\begin{table}[t!]
\caption{Leave one site out reproducbility on real dataset($k_2=4$).\label{tbl:leave_repro_h1}}
\centering
        \begin{tabular}{ lcccc }
 %\hline
 %\multicolumn{5}{|c|}{Comparison by varying number of components}\\
 \toprule
 {Method} & {$k_1=10$} & {$k_1=15$}  & {$k_1=20$} & {$k_1=25$} \\
 \midrule
hSCP & $0.652\pm0.038$   & $0.618\pm0.041$   &  $0.592\pm0.033$ &   $0.571\pm0.035$\\
ComBat hSCP & $0.614\pm0.042$   & $0.594\pm0.035$   &  $0.542\pm0.041$ &   $0.528\pm0.039$\\
Adv. hSCP & $0.656\pm0.035$ &  $0.629\pm0.039$  &  $0.601\pm0.035$ &    $0.584\pm0.034$ \\
rshSCP & $0.712\pm0.034$ &   $0.701\pm0.036$ &    $0.676\pm0.038$ &   $0.665\pm0.034$ \\
Adv. rshSCP & $\mathbf{0.716\pm0.032}$ & $\mathbf{0.709\pm0.031}$ & $\mathbf{0.688\pm0.034}$ &  $\mathbf{0.671\pm0.033}$ \\
 \bottomrule
\end{tabular}
\end{table} 
\begin{table}[t!]
\caption{Mean absolute error ($k_2=4$)\label{tbl:mae_h1}}
\centering
        \begin{tabular}{ lcccc }
 %\hline
 %\multicolumn{5}{|c|}{Comparison by varying number of components}\\
 \toprule
 {Method} & {$k_1=10$} & {$k_1=15$}  & {$k_1=20$} & {$k_1=25$} \\
 \midrule
hSCP & $6.490\pm 1.485$   & $6.468\pm 1.442$   &  $6.425\pm 1.412$ &   $6.414\pm 1.417$\\
%Adv. hSCP & $9\pm0.028$ &  $9\pm0.021$  &  $9\pm0.023$ &    $9\pm0.019$ \\
%rshSCP & $9\pm0.024$ &   $9\pm0.021$ &    $9\pm0.014$ &   $9\pm0.018$ \\
Adv. rshSCP & $6.494\pm 1.501$ & $6.467\pm 1.475$ & $6.432\pm 1.483$ &  $6.409\pm 1.470$ \\
 \bottomrule
\end{tabular}
\end{table} % 
\vspace{-\baselineskip}
\paragraph{Site prediction.} We performed the same experiment under the same settings as mentioned in the previous section to check $\mathbf{\Lambda}$ has reduced predictive power to predict the site. Using SVM, our model leads to a decrease in average cross-validation accuracy from $51\%$ to $32\%$. Using the first neural network architecture defined in section \ref{sec:simu}, the cross-validation accuracy for hSCP model is $59.3\%$ and for the rshSCP is $33.6\%$. Using the second architecture, the cross-validation accuracy for hSCP model is $58.7\%$ and for the rshSCP is $33.4\%$. This suggests that our model can reduce the prediction capability to predict site.

\paragraph{Age prediction}
We used subject specific information ($\mathbf{\Lambda}$) to predict age of each subject as the metric to check if our method is able to preserve age related biological variation. Table \ref{tbl:mae_h1} shows $10$ fold cross validation mean absolute error (MAE) using random forest . From the table, we can see that the proposed method has comparable performance as the hSCP suggesting that it preserves age related biological variance. More details about the experiment are given in Appendix \ref{sec:supp_real}.

\iffalse
\subsection{Analysis of components}
\label{sec:analysis}
\input{tables/mae_h1}
 \begin{figure}
  \centering
  \begin{minipage}{.48\linewidth}
    \centering
    \subcaptionbox{First top left}
      {\includegraphics[width=\linewidth]{figures/vanilla_DAN.png}}

    \subcaptionbox{Second top left}
      {\includegraphics[width=\linewidth]{figures/7_old.png}}

    \caption{Components estimated using hSCP}
  \end{minipage}\quad
  \begin{minipage}{.48\linewidth}
    \centering
    \subcaptionbox{First top right}
      {\includegraphics[width=\linewidth]{figures/DMN_DAN.png}}

    \subcaptionbox{Second top right}
      {\includegraphics[width=\linewidth]{figures/7_new.png}}

    \caption{Components estimated using rshSCP}
  \end{minipage}
\label{fig:comp_camp} 
\caption{Nodes  with  red  and  blue  color  arecorrelated among themselves, but are anitcorrelated with each other. Note that blue color does not need to be necessarily associatewith positive or negative correlation because the colors can be flipped without affecting the solution. }
\end{figure}
\fi
\subsection{Analysis of components}
\label{sec:analysis}

 \begin{figure}
  \centering
   \begin{minipage}{.49\linewidth}
   \centering
         \subcaptionbox{Component $5:$ $\rho = -0.02$, $-\ln(p) = 1.57$}
         {\makebox[0.87\linewidth][c]{\includegraphics[width=.65\linewidth]
    {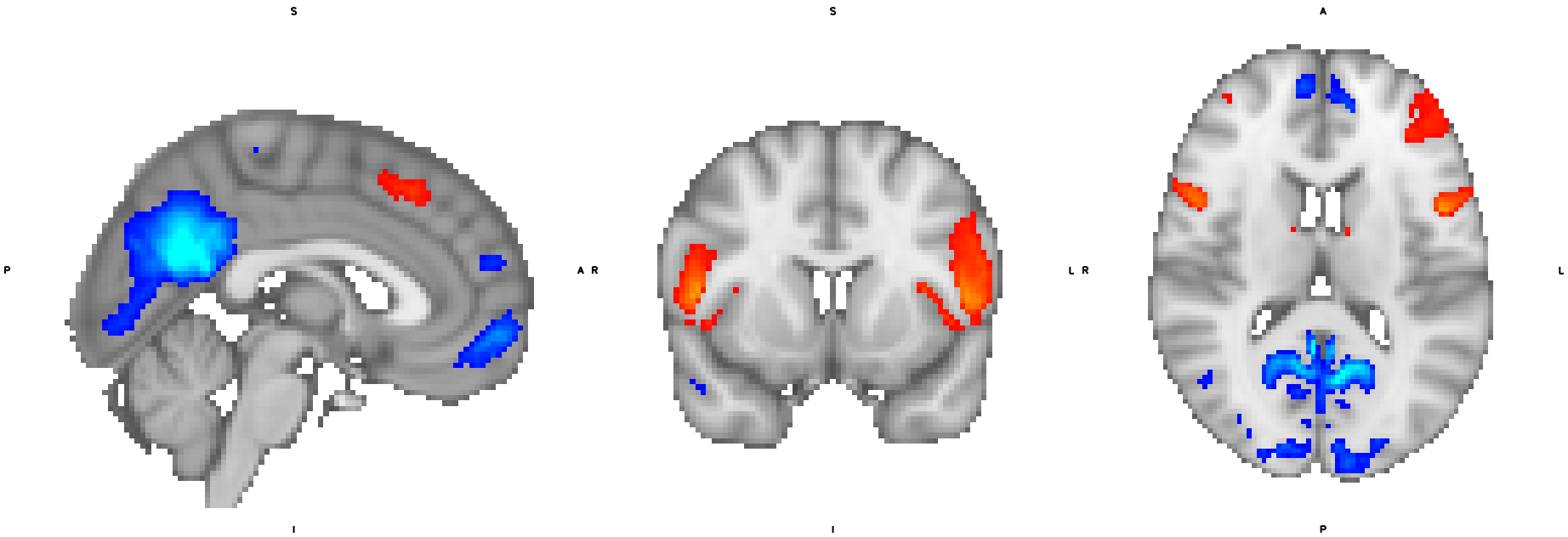}}}
      %{\includegraphics[width=0.87\linewidth]{figures/vanilla_DAN.png}}
        \end{minipage}
\begin{minipage}{.49\linewidth}
\centering
 \subcaptionbox{Component $5:$ $\rho = -0.13$, $-\ln(p) = 28.6$}
          {\makebox[0.87\linewidth][c]{\includegraphics[width=.65\linewidth]
    {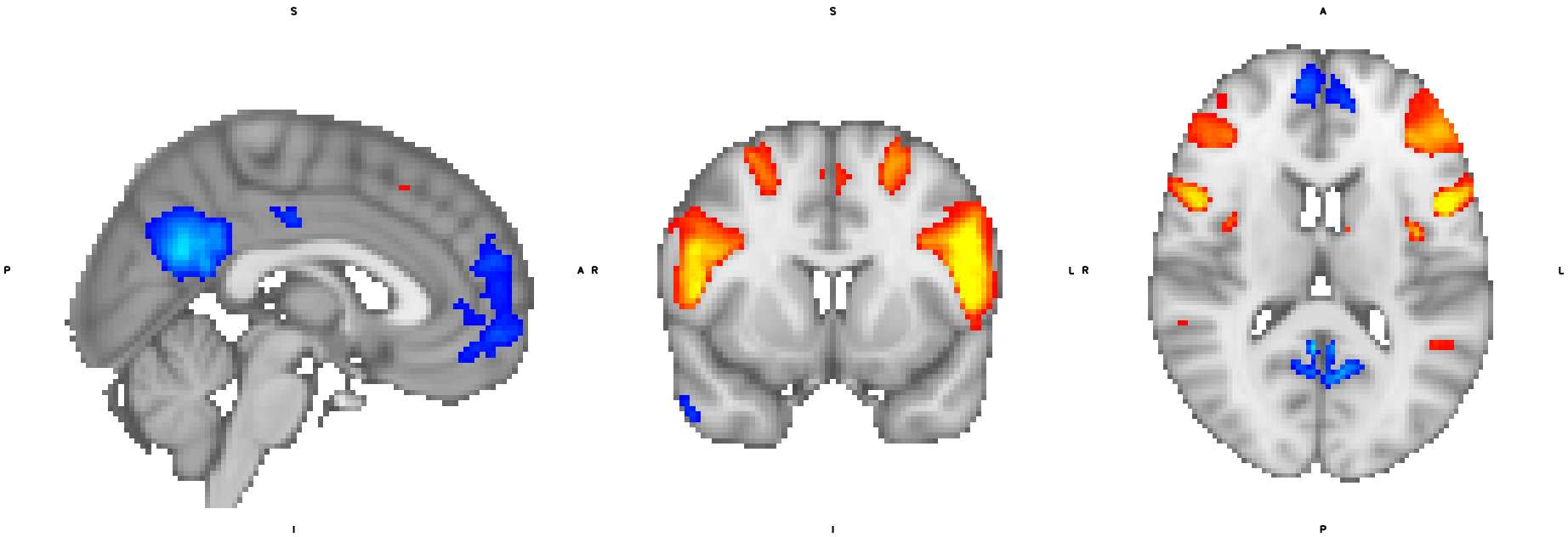}}}
      %{\includegraphics[width=.87\linewidth]{figures/DMN_DAN.png}}
  \end{minipage}
  \begin{minipage}{.49\linewidth}
  \centering
   \subcaptionbox{Component $6:$ $\rho = -0.07$, $-\ln(p) = 9.32$}
   {\makebox[0.87\linewidth][c]{\includegraphics[width=.65\linewidth]
    {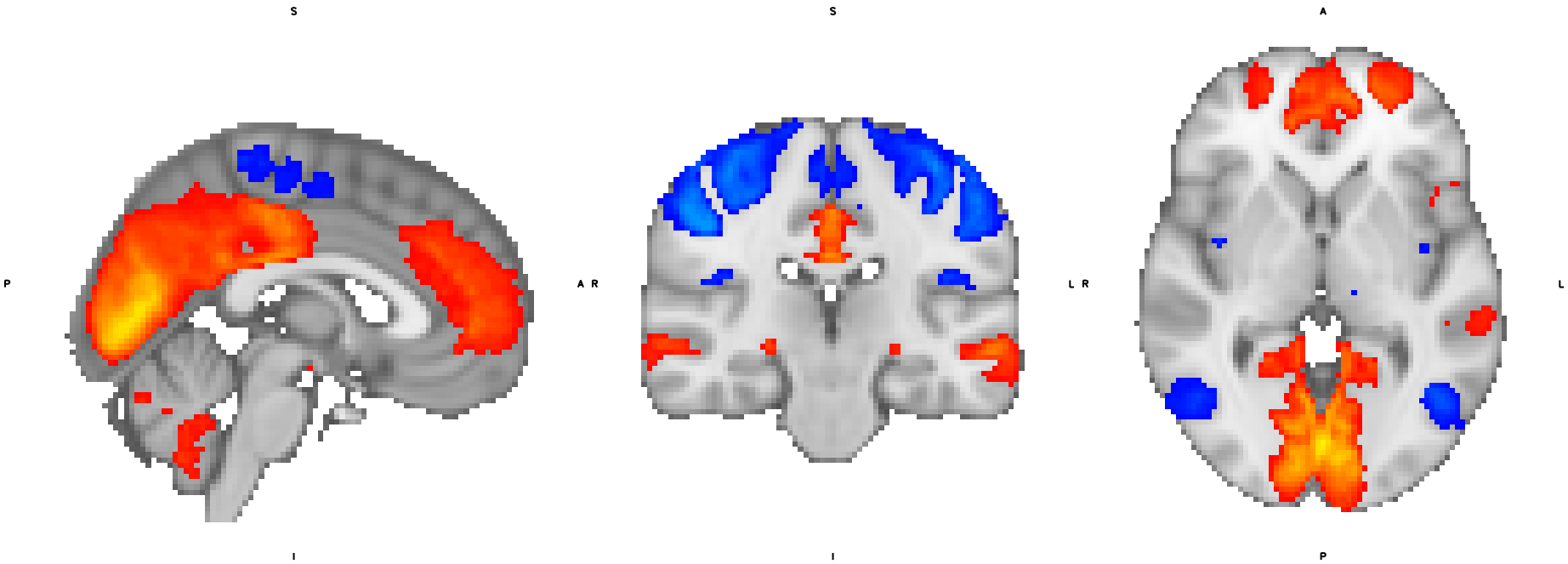}}}
      %{\includegraphics[width=.87\linewidth]{figures/7_old.png}}
        \end{minipage}
\begin{minipage}{.49\linewidth}
\centering
 \subcaptionbox{Component $6:$ $\rho = -0.05$, $-\ln(p) = 5.27$}
 {\makebox[0.87\linewidth][c]{\includegraphics[width=.65\linewidth]
    {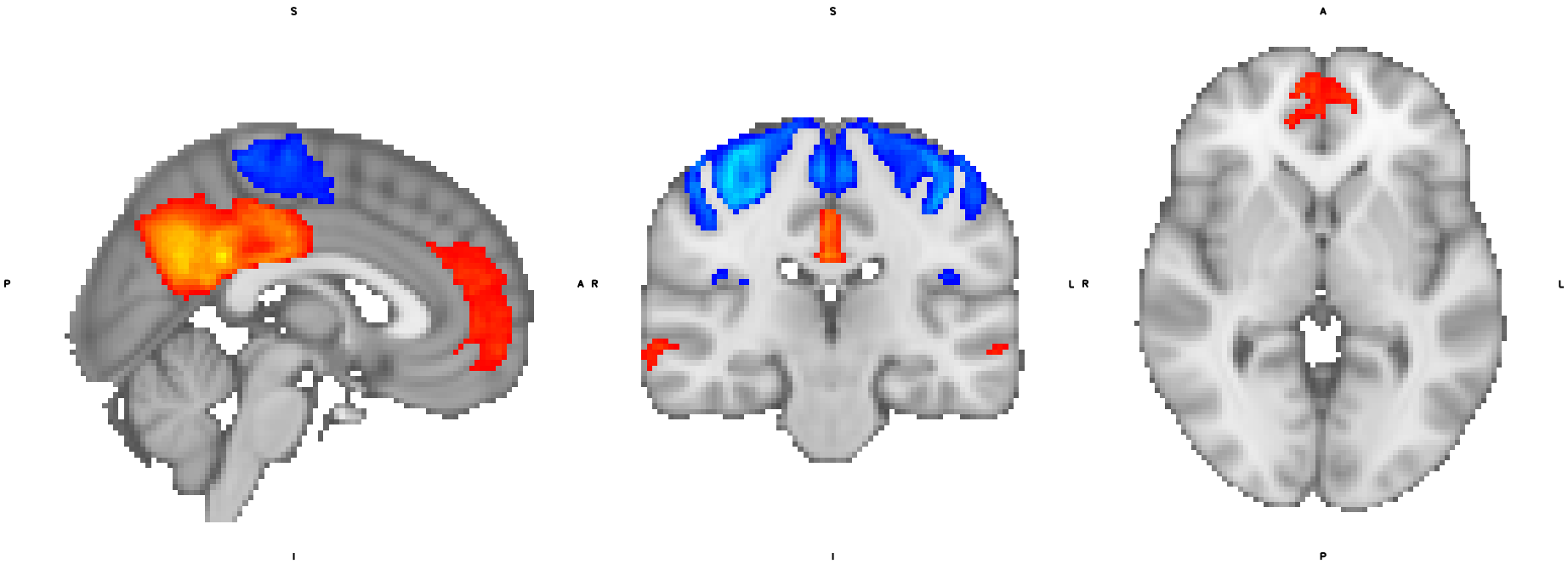}}}
      %{\includegraphics[width=.87\linewidth]{figures/7_new.png}}
  \end{minipage}
\caption{Left column ((a) \& (c)) displays the components estimated using hSCP and right column ((b) \& (d)) displays the components estimated using rshSCP. Red and blue regions are anti-correlated with each other but are correlated among themselves. The colors are not associated with negative or positive correlation since they can be swapped without affecting the final inference.\label{fig:comp_camp} }
\end{figure}
\iffalse
 \begin{figure}
  \centering
   \begin{minipage}{.71\linewidth}
   \centering
         \subcaptionbox{\RNum{2}: $\rho = -0.05$, $-\ln(p) = 5.34$}
      {\includegraphics[width=0.5\linewidth]{figures/hir1.png}}
        \end{minipage}
  \begin{minipage}{.49\linewidth}
  \centering
   \subcaptionbox{$2:$ $\rho = -0.02$, $-\ln(p) = 0.96$}
      {\includegraphics[width=.7\linewidth]{figures/3.png}}
        \end{minipage}
\begin{minipage}{.49\linewidth}
\centering
 \subcaptionbox{$7:$ $\rho = -0.35$, $-\ln(p) = 2.46$}
      {\includegraphics[width=.7\linewidth]{figures/8.png}}
  \end{minipage}
\caption{Hierarhical component (a)   displays the components estimated using hSCP and right column ((b) \& (d)) displays the similar components estimated using rshSCP. Red and blue regions are anti-correlated with each other but are correlated among themselves. The colors are not associated with negative or positive correlation since they can be swapped without affecting the final inference.\label{fig:hir} }
\end{figure}
\fi
\begin{figure}[t]
\centering
\begin{forest}
  styleA/.style={top color=white, bottom color=white},
  styleB/.style={%
    top color=white,
    bottom color=red!20,
    delay={%
      content/.wrap value={##1\\{\includegraphics[scale=.5]{hand}}}
    }
  },
for tree={
    edge path={\noexpand\path[\forestoption{edge}] (\forestOve{\forestove{@parent}}{name}.parent anchor) -- +(0,-16pt)-| (\forestove{name}.child anchor)\forestoption{edge label};}
},
  forked edges, s sep=20mm
[\RNum{2}{ \includegraphics[trim={1cm 1cm 1cm 1cm},clip,scale=0.09]{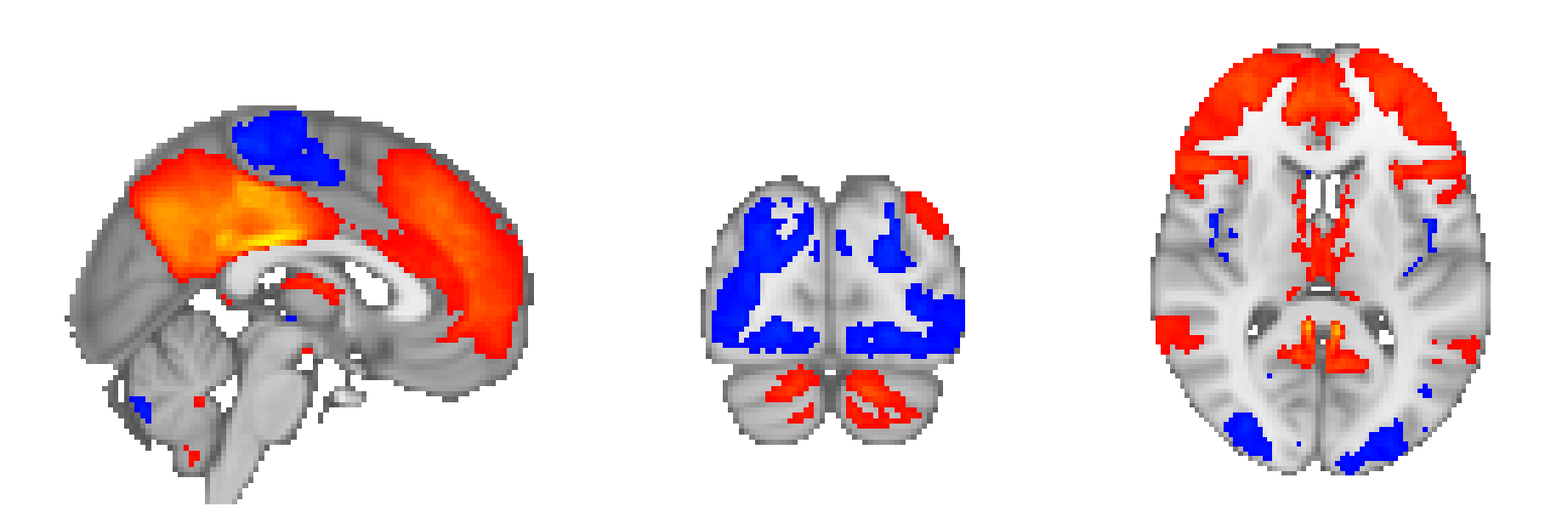}}
    [2{ \includegraphics[trim={0cm 1cm 1cm 0cm},clip,scale=0.09]{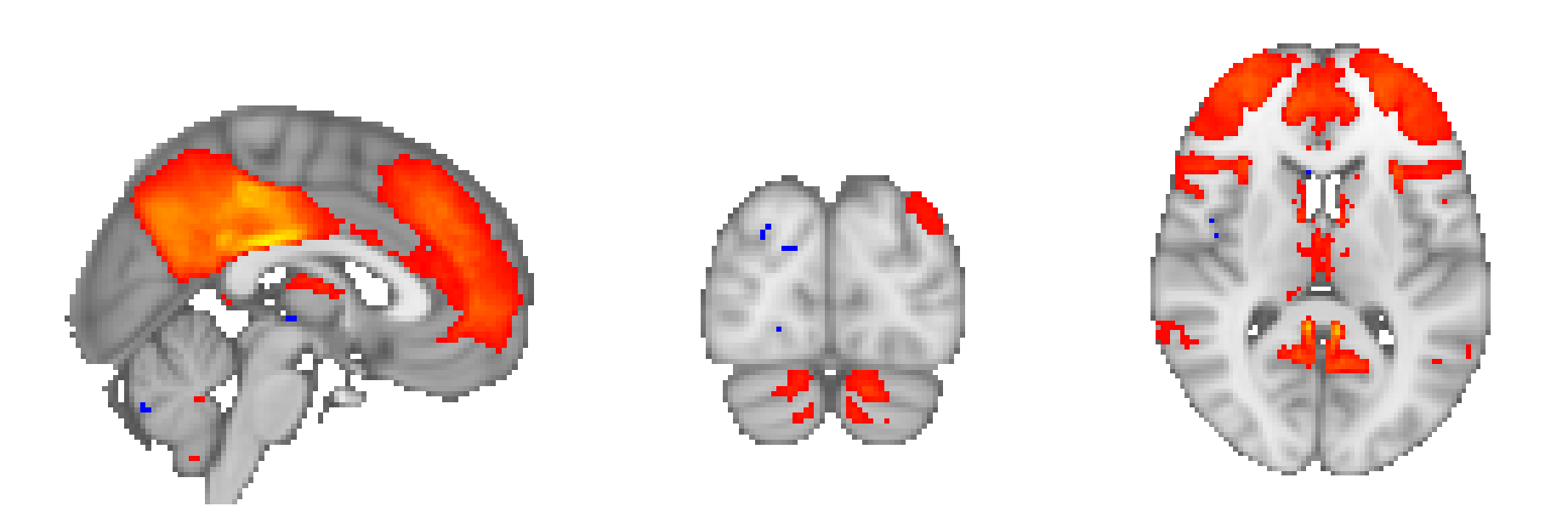}}]
    [7{ \includegraphics[trim={0cm 1cm 1cm 0cm},clip,scale=0.09]{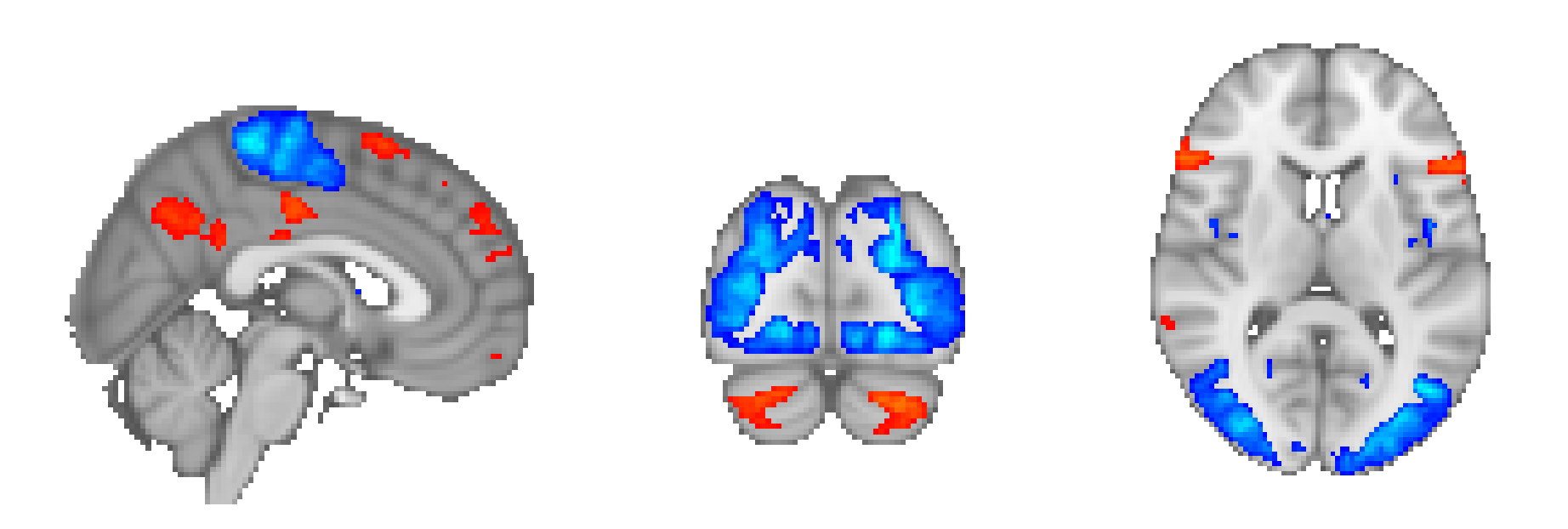}}]
     ]
  ]
\end{forest}
\caption{One of the hierarchical components derived from rshSCP comprising of component $2$ and $7$ at fine scale and component \RNum{2} at coarse scale. \label{fig:hirr}}
\end{figure}%A robust method should be able to remove most or possibly all non-biological variability caused by site and scanner while retaining biological variability.
A robust method should be able to reduce non-biological variability caused by site and scanner while retaining biological variability. In this study, we look at brain aging-related associations and leave analysis with other variables for future work. We also discuss the difference between the components with and reduced site effects. We selected the subjects with age greater than $60$ to find an association between brain aging and the components derived from hSCP and rshSCP. The total number of subjects having an age greater than $60$ is $2746$. We first computed Spearman correlation between subject-specific information ($\mathbf{\Lambda}$) and their respective age and used $0.05$ as the significance level for the hypothesis test of no correlation against the alternative hypothesis of a nonzero correlation. We derived $10$ fine-scale ($1-10$) and $4$ coarse-scale components (\RNum{1}-\RNum{4}) because of the high split sample reproducibility and easier interpretation of each component. The correlation and $p$-values are displayed in Table \ref{tbl:corr_pval_k1} and \ref{tbl:corr_pval_k2} in Appendix \ref{sec:supp_real} for the hSCP and rshSCP. 

We first compare components from the two methods. Figure \ref{fig:comp_camp} shows the components derived from hSCP and the proposed method. The first row of the figure displays the components with anti-correlation between Default Mode Network (DMN) and Dorsal Attention Network (DAN). The component derived using hSCP has a part of the visual area positively associated with DMN, but the opposite is true, as shown by the previous sparse connectivity patterns \cite{eavani2015identifying}. On the other hand, the component with reduced site effects is cleaner since it does not include that relation. This component has a negative correlation with age which has been previously shown in resting-state fMRI and task-based fMRI \cite{spreng2016attenuated}. The magnitude of anti-correlation has been connected to individual differences in task performances in healthy young adults \cite{keller2015resting}. However, in the case of older adults, the behavioral implications of reduced anti-correlation remain unclear. The second row of the Figure \ref{fig:comp_camp} displays another set of components for comparison. The components stores information about the anti-correlation between DMN and sensorimotor, which aligns with the previous literature \cite{karahanouglu2015transient}. But the addition of a positive correlation of DMN with visual areas will cause misleading inference since it contradicts the previous SCPs and studies. Hence making an inference without removing the site effects can be misleading.  Discussion of the association between other relevant components and aging is in Appendix \ref{sec:supp_real} showing biological interpretability of the components in brain aging. From the results, we can see that there is an increase (or decrease in anti-correlation) in connectivity between different networks in the aging brain. This suggests that there is a reorganization of the aging brain aligning with the previous findings \cite{damoiseaux2017effects}. This can serve as a base to explore rshSCP as a biomarker of neurodegenerative diseases. 

Figure \ref{fig:hirr} displays one of the hierarchical components with coarse-scale component storing relation between different fine-scale components comprising DMN, sensorimotor, and visual areas, previously studied by \cite{karahanouglu2015transient}. These findings give evidence that even after removal site effects, the components can have a meaningful interpretation. The results indicate that our approach can extract robust informative patterns without using traditional seed-based methods that are dependent on the knowledge of the seed region of interest.

\section{Conclusion}
\label{sec:conclusion}
In this work, we have presented a method for estimating site effects in hSCP. We formulated the problem as a minimax non-convex optimization problem and solved it using AMSgrad. We also propose a simple initialization procedure to make the optimization procedure deterministic and improve the performance on an average on a simulated and real dataset. Experimentally, using a simulated dataset, we showed that our method accurately estimates the ground truth compared to the vanilla method with better reproducibility. On the real dataset, we show that the proposed method can capture components with a better split sample and leave one site out reproducibility without losing biological interpretability and information. We also show that without removal of site effects, we can have a noisy estimate of sparse components resulting in misleading downstream analysis.

Below we mention some directions for future research. First, it would be interesting to consider the framework for the analysis of task-induced activity to investigate the extent of site effect and corrections on underlying networks activated by the task. Second, one could look at the changes in the associations of hSCPs with various clinical variables such as Mini-mental score, Digit Span Forward score, etc., after removing site effects. Third, we can also look low-dimensional modeling of $\mathbf{V}$ along with sparse constraints which has been used several robust matrix factorization problems. Since we have only shown age related biological preservation, future studies will focus on whether the proposed method preserves components associated with other demographic, clinical phenotypes, and pathological biomarkers.  

There are few weaknesses of our proposed model, which also adds directions for future work. First, our method only captures linear site effects, it would be interesting to see if explicitly capturing non-linear site effects can improve the performance of the model. Second, the result of the optimization algorithm depends on the initialization procedure, which has been shown to perform well on the simulated dataset and real dataset but can be sub-optimal.

\normalsize
\section{Broader Impact}
In this work, we provide a new method to diminish site effects in hSCPs and robustly estimate the components in multi-center studies. The formulation used in the method is not limited to hSCP. It can be easily extended to various matrix factorization approaches such as Independent Component Analysis, Non-negative Matrix Factorization, Dictionary Learning, etc., to improve the reproducibility of functional networks/components. Our work not only has broader applicability in terms of methods used for estimation of components but also to different types of neuroscience data, which includes EEG, MEG, etc.

Meanwhile, it should be noted that any unsupervised machine learning model has risk associated with it in terms of accuracy of the model; our work is no departure from this. It can lead to a negative impact since the reproducibility of the components is not $100\%$, which can lead to misleading clinical and biological interpretability of some of the components. This can be prevented by analyzing the components having high reproducibility as shown in section \ref{sec:real_dataset} and \ref{sec:analysis}.

\footnotesize
\bibliography{main.bib}
\bibliographystyle{unsrtnat}

\normalsize

\newpage
\appendix

\section{Related Work: Extended}
\label{sec:rel_work}
\subsection{Methods for Tackling Site Effects}
One of the first investigations of batch effects in rs-fMRI was performed by \citet{olivetti2012adhd} using extremely randomized trees along with dissimilarity representation. One of the common methods to remove site effects is the harmonization of data. Harmonization of fMRI data especially derived measures, is very nascent, even though it is much needed with the growing number of multi-site data sets \cite{adhikari2019resting}. Recently, \citet{yu2018statistical} used ComBat harmonization \cite{johnson2007adjusting} to remove site effects in connectivity matrices. However, ComBat and its variants such as ComBat-GAM \cite{pomponio2020harmonization} can not be directly applied to connectivity matrices since it can destroy the structure of the connectivity matrix and semi definiteness of the connectivity matrix. A similar difficulty arises when applying ComBat based harmonization to other structured data. Another approach is not to remove site effects, but directly use site information for downstream analysis such as age prediction, finding associations with various clinical variables, etc. \citet{kia2020hierarchical,bayer2021accommodating} used normative modeling for the age prediction task while keeping the site as one of the predictors. One limitation of the method is that without removing the site effects, the biomarkers can not be used for downstream analysis by the clinician, psychiatrist, etc., directly, which is one of the goals of the hSCP. 

Recent work by \citet{vega2018finding} analyzed the impact of covariate analysis, z-score normalization, and whitening on batch effects. Domain adaption has also been introduced in removing batch effects in rs-fMRI data. Domain adaption techniques aim to learn from multiple sources and generalize the model to perform well on a new related target site. Extensive work has been done on unsupervised domain adaptation approaches \cite{gholami2020unsupervised,zhao2019multi}. Several methods have been introduced for domain adaptation such as Multi-source Domain Adversarial Networks \cite{zhao2018adversarial}, Multi-Domain Matching Networks \cite{li2018extracting}, Moment Matching \cite{peng2019moment}, etc. Readers can refer to the detailed survey by \citet{zhao2020multi}. In multi-site fMRI data, \citet{wang2019identifying} introduced a low-rank domain to remove batch effects. Other recent approached include transport-based joint distribution alignment \cite{zhang2020transport} and federated learning \cite{li2020multi} for fMRI data.

\subsection{Hierarchical Decomposition}
The hierarchical organization has been observed in large-scale computer architectures \cite{ozaktas1992paradigms}, communication systems \cite{AKYILDIZ2005445}, and social networks \cite{nickel2017poincare}. Such an organization provides a unique solution to balancing information within a group at a single scale and between groups at multiple scales. It also promotes optimal and efficient information processing and transmission in real-world information processing systems \cite{kinouchi2006optimal}. The hierarchical organization is also seen in natural information processing systems such as the human brain, where this organization is present both in space \cite{bassett2010efficient} and time \cite{chaudhuri2014diversity}. Our understanding of hierarchical network structure in the brain is limited, despite having fundamental importance, due to its complex nature. 

Several different fields have evolved with a particular method to analyze brain organization. One such widely used method is community detection. Several interesting multi-scale community detection methods have been developed for estimating the underlying hierarchical organization of human brain connectivity \cite{ferrarini2009hierarchical,al2018tensor,akiki2019determining,ashourvan2019multi}. The major limitations of the community detection approaches are one or more than one of the following: 1) the assumption of independent components, 2) not capturing heterogeneity in the data, and 3) inability to detect weights while estimating links. In addition, community detection approaches remove negative edge links before the analysis as they treat them repulsion. Whereas, in the fMRI, negative links carry essential that can play an essential role in analyzing neuropsychiatric disorders \cite{fitzpatrick2007associations} has a substantial physiological basis \cite{zhan2017significance,fox2005human}. Another set of methods that have recently evolved is based on deep learning approaches \cite{hu2018latent,zhang2020hierarchical,dong2019modeling}. These methods have reported the meaningful hierarchical temporal organization of fMRI time series in the task-evoked fMRI data. Still, there are several disadvantages: 1) non existence of positively and negatively correlated nodes in a component, 2) inability to capture heterogeneity in the data, and 3) ``black box'' results lacking explainability mainly due to non-linearity in the hierarchical associations. 
\section{Algorithm}
\label{sec:alg}
\subsection{Gradient Calculations}
In this section, we define gradients used for alternating gradient descent. Let
\begin{align*}
\mathbf{\tilde{W}}_0 &= \mathbf{W}_0 = \mathbf{I}_P, \qquad
\mathbf{Y}_r = \prod_{j=0}^{r}\mathbf{W}_j, \qquad \mathbf{\tilde{Y}}_r = \prod_{j=0}^{r}\mathbf{\tilde{W}}_j, \\ 
\mathbf{T}_{m,n}^r &= (\prod_{j=1}^{m-r}\mathbf{W}_j)\mathbf{\Lambda}^n_{m-r}(\prod_{j=1}^{m-r}\mathbf{W}_j)^\top, \qquad \mathbf{\tilde{T}}_{m,n}^r = (\prod_{j=1}^{m-r}\mathbf{\tilde{W}}_j)\mathbf{\Lambda}^n_{m-r}(\prod_{j=1}^{m-r}\mathbf{\tilde{W}}_j)^\top, \\
\mathbf{X}^n_r &= \mathbf{\Theta}^n - \mathbf{U}^s_r \mathbf{V}_r, \qquad \mathbf{Z}^n_r = \mathbf{\Theta}^n - (\prod_{j=1}^{r}\mathbf{W}_j)\mathbf{\Lambda}_r^n(\prod_{n=1}^{r}\mathbf{W}_n)^\top, 
\end{align*}
where $n \in \mathcal{I}_s$, $\mathbf{X}^n_r$ stores the information after removing site effects from $\mathbf{\Theta}^n$ and $\mathbf{Z}^n_r$ stores the information after removing subject-wise and shared component information at the $r$th level. We first define gradient for updating adversarial perturbations $\mathbf{\tilde{W}_r}$. The gradient of classifier loss with respect to $\mathcal{D}$ is calculated using automatic differentiation provided by \textsc{matlab}. The objective function is $F = \alpha\| \mathbf{\hat{W}}_r - \mathbf{{W}}_r\|_F^2 + H(\mathcal{\tilde{W}},\mathcal{D},\mathcal{P}) $ and gradient with respect to $\mathbf{\tilde{W}_r}$ will be
\begin{align*}
    \frac{F}{\partial \mathbf{\tilde{W}_r}} & = 2\alpha(\mathbf{\hat{W}}_r - \mathbf{{W}}_r) + \frac{\partial H(\mathcal{\tilde{W}},\mathcal{D},\mathcal{C})}{\partial \mathbf{\tilde{W}}_r} \\ & = 2\alpha(\mathbf{\hat{W}}_r - \mathbf{{W}}_r)+  \sum_{n=1}^N\sum_{j=r}^{K} \bigl(-4\mathbf{\tilde{Y}}_{r-1}^\top\mathbf{\Gamma}^n\mathbf{\tilde{Y}}_{r-1}\mathbf{\tilde{W}}_r\mathbf{\tilde{T}}_{j,n}^r \\& +   4\mathbf{\tilde{Y}}_{r-1}^\top\mathbf{\tilde{Y}}_{r-1}\mathbf{\tilde{W}}_r\mathbf{\tilde{T}}_{j,n}^r\mathbf{\tilde{W}}_r^\top\mathbf{\tilde{Y}}_{r-1}^\top\mathbf{\tilde{Y}}_{r-1}\mathbf{\tilde{W}}_r\mathbf{\tilde{T}}_{j,n}^r \bigr). 
\end{align*}
We now define gradients for updating model parameters. The gradient of objective function $J$ with respect to $\mathbf{\Lambda}_r^i$ is:
\begin{align*}
\frac{\partial J}{\partial \mathbf{\Lambda}_r^i} & = \frac{\partial H(\mathcal{\tilde{W}},\mathcal{D},\mathcal{C})}{\partial \mathbf{\Lambda}_r^i} + \beta \frac{\partial H(\mathcal{{W}},\mathcal{D},\mathcal{C})}{\partial \mathbf{\Lambda}_r^i}  + \gamma \frac{\partial \mathcal{L}(\zeta, \mathcal{D}, \mathbf{y})}{\partial \mathbf{\Lambda}_r^i} 
\\ & \hspace{-2em} = \left[(-2\mathbf{\tilde{Y}}_r^\top\mathbf{X}_r^i\mathbf{\tilde{Y}}_r + 2\mathbf{\tilde{Y}}_r^\top\mathbf{\tilde{Y}}_r\mathbf{\Lambda}_r^i\mathbf{\tilde{Y}}_r^\top\mathbf{\tilde{Y}}_r)  + \beta (-2\mathbf{Y}_r^\top\mathbf{X}_r^i\mathbf{Y}_r + 2\mathbf{Y}_r^\top\mathbf{Y}_r\mathbf{\Lambda}_r^i\mathbf{Y}_r^\top\mathbf{Y}_r)\right] \circ \mathbf{I}_{k_r} + \gamma \mathbf{F},
\end{align*}
where is $\mathbf{F}$ i.e $\frac{\partial \mathcal{L}(\zeta, \mathcal{D}, \mathbf{y})}{\partial \mathbf{\Lambda}_r^i}$ is calcualted using automatic differentiation toolbox in \textsc{MATLAB}.
The gradient of $J$ with respect to $\mathbf{W}_r$ is:
\begin{align*}
\frac{\partial J}{\partial \mathbf{W}_r} &= \frac{\partial H(\mathcal{{W}},\mathcal{D},\mathcal{C})}{\partial \mathbf{W}_r}\\ &= \sum_{n=1}^N\sum_{j=r}^{K} -4\mathbf{Y}_{r-1}^\top\mathbf{X}_n\mathbf{Y}_{r-1}\mathbf{W}_r\mathbf{T}_{j,n}^r +   4\mathbf{Y}_{r-1}^\top\mathbf{Y}_{r-1}\mathbf{W}_r\mathbf{T}_{j,n}^r\mathbf{W}_r^\top\mathbf{Y}_{r-1}^\top\mathbf{Y}_{r-1}\mathbf{W}_r\mathbf{T}_{j,n}^r.  
\end{align*}
The gradient $J$ with respect to$\mathbf{U}^s$ and $\mathbf{V}$ are:
\begin{align*}
    \frac{\partial J}{\partial \mathbf{U}^s} = \left(\sum_{n=\mathcal{I}_s} \left(\mathbf{Z}^n - \mathbf{U}^s \mathbf{V} \right)\mathbf{V}^\top \right) \circ \mathbf{I}_{p}
\end{align*}
\begin{align*}
    \frac{\partial J}{\partial \mathbf{V}} = \sum_{s=1}^S\sum_{n \in \mathcal{I}_s} \mathbf{U}^s \left(\mathbf{Z}^n - \mathbf{U}^s \mathbf{V} \right)
\end{align*}
 \begin{algorithm}[t]
 \caption{rshSCP}\label{alg:site_HSCP}
\begin{algorithmic}[1]
\State \textbf{Input:} Data $\mathcal{C}$, number of connectivity patterns $k_1$, \ldots, $k_K$ and sparsity $\tau_1$, \ldots, $\tau_K$ at different level, hyperparameters $\alpha$, $\beta$, $\gamma$ and $\mu$. 
\State Initialize $\mathcal{W}$ and $\mathcal{D}$ using $\svdinit$
\State Initialize $\mathcal{U}$ and $\mathbf{V}$ using equation \ref{eq:init}
\Repeat 
\For{$r=1$ {\bfseries to} $K$} 
\If{Starting criterion is met}
\State \textit{Update adversarial perturbations}
\State $\mathbf{\hat{W}}_r \leftarrow \descent(\mathbf{\hat{W}}_r, \alpha)$
\State $\mathbf{W}_r \leftarrow \descent(\mathbf{W}_r)$
\EndIf
%\State $W_l \leftarrow \left( W_l - \eta \nabla_{W_l}H \right) $
\If{$r==1$}
   \State $\mathbf{W}_r \leftarrow \proj_1(\mathbf{W}_r,\tau_r) $
   \Else{}
\State $\mathbf{W}_r \leftarrow \proj_2(\mathbf{W}_r) $
   \EndIf
\For{$n= 1,..,N$} 
\State $\mathbf{\Lambda}_r^n \leftarrow \descent(\mathbf{\Lambda}_r^n, \beta, \gamma)$
\State $\mathbf{\Lambda}_r^n \leftarrow \proj_2(\mathbf{\Lambda}_r^n) $
\EndFor
\EndFor
\For{$s=1$ {\bfseries to} $S$}
\State $\mathbf{U}^s \leftarrow \descent(\mathbf{U}^s)$
\EndFor
\State $\mathbf{V} \leftarrow \descent(\mathbf{V})$
\State $\mathbf{V} \leftarrow \proj_3(\mathbf{V},\mu)$
\Until{Stopping criterion is reached}
\State \textbf{Output:} $\mathcal{W}$ and $\mathcal{L}$
\end{algorithmic}
\end{algorithm} 
\subsection{Alternating Minimization}
Algorithm~\ref{alg:site_HSCP} describes the complete alternating minimization procedure. $\mathcal{W}$ and $\mathcal{D}$ are initialized using $\svdinit$ algorithm \cite{sahoo2020hierarchical}[Algorithm 2], and $\mathcal{U}$ and $\mathbf{V}$ according to the equation \ref{eq:init}. $\proj_1(\mathbf{W},\tau)$ operator is used for projecting each column of $\mathbf{W}$ into the intersection of $L_1$ and $L_\infty$ ball \cite{podosinnikova2013robust}, $\proj_2$ operator is used for projecting a matrix onto $\mathbb{R}_{+}$ by zeroing out all then negative values in the matrix, and $\proj_3$ operator is used for projection onto $L_1$ ball. We use AMSGrad \cite{reddi2019convergence} denoted as $\descent$ in the algorithm with gradients defined in the previous section for performing gradient descent for all the variables. $\beta_1$ and $\beta_2$ are kept to be $0.9$ and $0.999$ in AMSGrad based on the experiments of \cite{sahoo2020hierarchical,sahoo2021extraction}. We start adversarial training only after the convergence of all the variables. We found that the algorithm uses $ 200$ iterations to reach convergence initially, as shown in Figure \ref{fig:convg}. The reason being that adversarial learning can start from an optimal point on which it can improve upon if there is overfitting. When the adversarial learning starts, first, the adversarial perturbations are generated by performing gradient descent on $\mathbf{\hat{W}}_r$, and then the model parameters are updated using gradient descent. This process is repeated until the convergence criteria is met.

\subsection{Convergence results}
\begin{figure}
   \centering
\begin{subfigure}{.45\textwidth}
\centering
     \includegraphics[width=.79\linewidth]{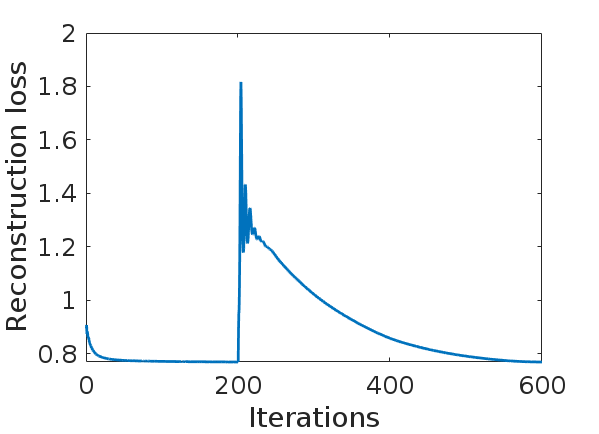}
    \caption{$k_1 = 10, k_2 = 4$}   
\end{subfigure}
 \begin{subfigure}{.45\textwidth}
 \centering
           \includegraphics[width=.79\linewidth]{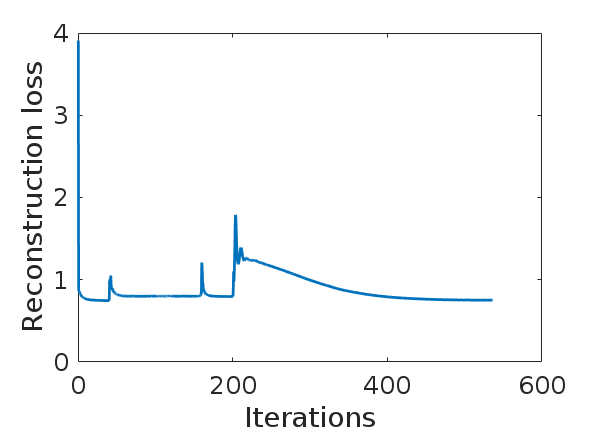}
    \caption{$k_1 = 10, k_2 = 6$}  
 \end{subfigure}
    \caption{Convergence of rshSCP algorithm using the complete dataset for different values of $k_2$. In the figure, for the first $200$ iterations, the algorithm converges without the adversarial perturbations. As the adversarial perturbations are introduced, the loss starts to oscillate where the adversarial perturbations force the algorithm to deviate from the optimal value. In defense, we minimize the objective function until convergence is reached. \label{fig:convg}}
\end{figure}

We empirically validate the convergence of Algorithm~\ref{alg:site_HSCP} using the reconstruction error:
\begin{align*}
    \frac{ \sum_{s=1}^S \sum_{n \in \mathcal{I}_s} \sum_{r=1}^{K}||\mathbf{\Theta}^i - (\prod_{j=1}^{r}\mathbf{W}_j)\mathbf{\Lambda}_r^i(\prod_{j=1}^{r}\mathbf{W}_j)^\top - \mathbf{U}^s \mathbf{V}||_F^2}{\sum_{n=1}^{N} \sum_{r=1}^{K}||\mathbf{\Theta}^i ||_F^2}.
\end{align*}
Figure \ref{fig:convg} shows the convergence of the algorithm on the complete dataset. 

\section{Data Preprocessing}
\label{sec:pre}
The pooled dataset included scans of participants with absence of any known diagnosis of a neurological or psychiatric disorder. FMRIB Software \cite{jenkinson2012fsl} is used for initial pre-processing as a part of the UK Biobank pipeline. The steps included the removal of the first five volumes, head movement correction using FSL's MCFLIRT \cite{jenkinson2012fsl}, global 4D mean intensity normalization, and temporal high-pass filtering ($>0.01$ Hz). 

After standard pre-processing steps, we applied FIX (FMRIB’s ICA-based Xnoiseifier) \cite{salimi2014automatic,griffanti2014ica} to remove structured artefacts. In the next step, functional images were co-registered to T1 using FLIRT with BBR as the cost function, and T1-weighted images were registered to the MNI152 template using FSL's FNIRT (non-linear registration). We projected the data into a lower-dimensional space by extracting a set of group Independent Components \cite{smith2014group} having dimension $100$ from individual subjects. These ICA maps can be considered ``parcellations'' but contain a continuous range of values and not binary masks. For a given IC map, the group IC spatial maps were mapped onto each subject's resting fMRI time series to derive one representative time series per IC component using Group Information Guided ICA(GIGICA) \cite{du2013group}. 

Quality control of the dataset is based on below metrics- 
\begin{enumerate}[leftmargin=*]
 \item Mean Relative (frame-wise) Displacement (MRD): We used MRD calculated by MCFLIRT to quantify head motion \cite{jenkinson1999measuring}. We set a threshold $0.2$mm.
    \item Time course Signal to Noise Ratio (tSNR): tSNR is an important metric for evaluating the ability of the fMRI acquisition to detect neural signal changes in the time series. It is is defined as the ratio of mean intensity and standard deviation across time within the evaluated Region of Interest \cite{triantafyllou2005comparison}. We excluded the subjects having temporal SNR less than $100$.
    \item Framewise Displacement (FD): It evaluates the head motion of each volume compared to the previous volume \cite{power2012spurious, jenkinson2002improved}. We set the threshold for FD to be $0.2$ mm \cite{power2012spurious,yan2013addressing}.  
\end{enumerate}

\section{Additional experiments}
\label{sec:supp_exp}

\subsection{Simualted Dataset}
\label{sec:supp_simu}
\begin{table}[t!]
\caption{5 fold cross validation accuracy (\%) on simulated dataset \label{tbl:site_accu} at one level.}
\centering
        \begin{tabular}{ lcccc }
 %\hline
 %\multicolumn{5}{|c|}{Comparison by varying number of components}\\
 \toprule
 {Method} & {$k_1=8$} & {$k_1=10$}  & {$k_1=12$} & {$k_1=14$} \\
 \midrule
hSCP & $97.3\pm 0.3$    & $98.1\pm 0.4$ &   $97.1\pm 0.3$ & $97.9 \pm 0.2$\\
Adv. rshSCP & ${65.5 \pm 0.6} $ & $67.2 \pm 0.5$  &  $67.5 \pm 0.5$  &  $68.1 \pm 0.7$\\
 \bottomrule
\end{tabular}
\end{table}
\paragraph{Site prediction.} To check if the estimated subject information ($\mathbf{\Lambda}$) has reduced predictive power to predict the site to which the subject belonged, we performed a $5$ fold cross-validation using SVM with RBF kernel. For prediction using SVM, we directly use the inbuilt function of MATLAB. Since we already used feed forward network, we decided to use a different model. Prediction results for site using $\mathbf{\Lambda}$ are displayed in Table \ref{tbl:site_accu}. We can see that there is an average drop of $20\%$ accuracy.

\paragraph{Reproducibility.} Table \ref{tbl:simu_repro_h1} and \ref{tbl:simu_repro_h2} shows the split sample reproducibility of different methods on two level simulated dataset.
\begin{table}[t!]
\caption{Reproducbility on simulated dataset ($k_2 = 4$). \label{tbl:simu_repro_h1}}
\centering
        \begin{tabular}{ lcccc }
 %\hline
 %\multicolumn{5}{|c|}{Comparison by varying number of components}\\
 \toprule
 {Method} & {$k_1=8$} & {$k_1=10   $}  & {$k_1=12$} & {$k_1=14$} \\
 \midrule
hSCP & $0.801\pm0.037$    & $0.805 \pm 0.042$ &   $0.787 \pm 0.041$ & $0.772 \pm 0.037$\\
ComBat hSCP & $0.776\pm0.041$    & $0.756 \pm 0.044$ &   $0.753 \pm 0.045$ & $0.745 \pm 0.038$\\
Adv. hSCP & $0.808 \pm 0.036$     & $0.824 \pm 0.034$ &  $0.799 \pm 0.030$  & $0.783 \pm 0.038 $ \\
rshSCP & $0.850 \pm 0.037$     & $0.853\pm 0.031$ &  ${0.839\pm 0.035}$  & $0.805\pm 0.034$ \\
Adv. rshSCP & $\mathbf{0.852 \pm 0.036}$     &$\mathbf{0.861 \pm 0.038}$  &  $\mathbf{0.842 \pm 0.043}$  &  $\mathbf{0.813 \pm 0.035}$\\
 \bottomrule
\end{tabular}
\end{table}

\begin{table}[t!]
\caption{Reproducbility on simulated dataset ($k_2=6$). \label{tbl:simu_repro_h2}}
\centering
        \begin{tabular}{ lcccc }
 %\hline
 %\multicolumn{5}{|c|}{Comparison by varying number of components}\\
 \toprule
 {Method} & {$k_1=8$} & {$k_1=10   $}  & {$k_1=12$} & {$k_1=14$} \\
 \midrule
hSCP & $0.786\pm0.041$    & $0.801 \pm 0.039$ &   $0.771 \pm 0.042$ & $0.769 \pm 0.038$\\
ComBat hSCP & $0.779\pm0.044$    & $0.770 \pm 0.041$ &   $0.742 \pm 0.040$ & $0.734 \pm 0.045$\\
Adv. hSCP & $0.793 \pm 0.036$     & $0.828 \pm 0.038$ &  $0.789 \pm 0.035$  & $0.762 \pm 0.036 $ \\
rshSCP & $0.833 \pm 0.038$     & $0.846\pm 0.031$ &  $0.831\pm 0.032$  & $0.795\pm 0.034$ \\
Adv. rshSCP & $ \mathbf{0.841 \pm 0.039}$     &$\mathbf{0.851 \pm 0.035}$  &  $\mathbf{0.835 \pm 0.033}$  &  $\mathbf{0.808 \pm 0.039}$\\
 \bottomrule
\end{tabular}
\end{table}

\begin{table}[t!]
\caption{Change in accuracy of rshSCP with sparsity parameter ($\mu$) of $\mathbf{V}$ on real dataset at one level. \label{tbl:v_k1}}
\centering
        \begin{tabular}{ lcccc }
 %\hline
 %\multicolumn{5}{|c|}{Comparison by varying number of components}\\
 \toprule
 $\mu$ & {$k_1=8$} & {$k_1=10$}  & {$k_1=12$} & {$k_1=14$} \\
 \midrule
$0.1$ & $0.799$   & $0.756$   &  $0.761$ &   $0.782$\\
$0.5$ & $0.873$ &  $0.865$  &  $0.843$ &    $0.867$ \\
$1$ & $0.801$ &   $0.789$ &    $0.767$ &   $0.754$ \\
 \bottomrule
\end{tabular}
\end{table}

\subsection{Real Dataset}
\label{sec:supp_real}
\begin{figure}
    \centering
    \includegraphics[width=0.5\textwidth]{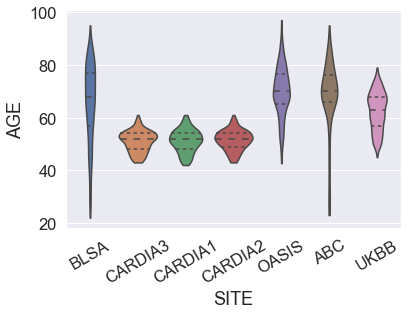}
    \caption{Violin plot of age for different sites.}
    \label{fig:violin}
\end{figure}
\paragraph{Reproducibility.} Table \ref{tbl:real_repro_h2} and Table \ref{tbl:leave_repro_h2} shows split sample reproducibility and leave one site out reproducibility respectively at two level for $k_2 = 6 $ and multiple values of $k_1$.

\begin{figure}[t!]
  \centering
    \begin{minipage}{.49\linewidth}
  \centering
   \subcaptionbox{Component $7:$ $\rho = -0.03$, $-\ln(p) = 2.46$}
   {\makebox[0.87\linewidth][c]{\includegraphics[width=.65\linewidth]
    {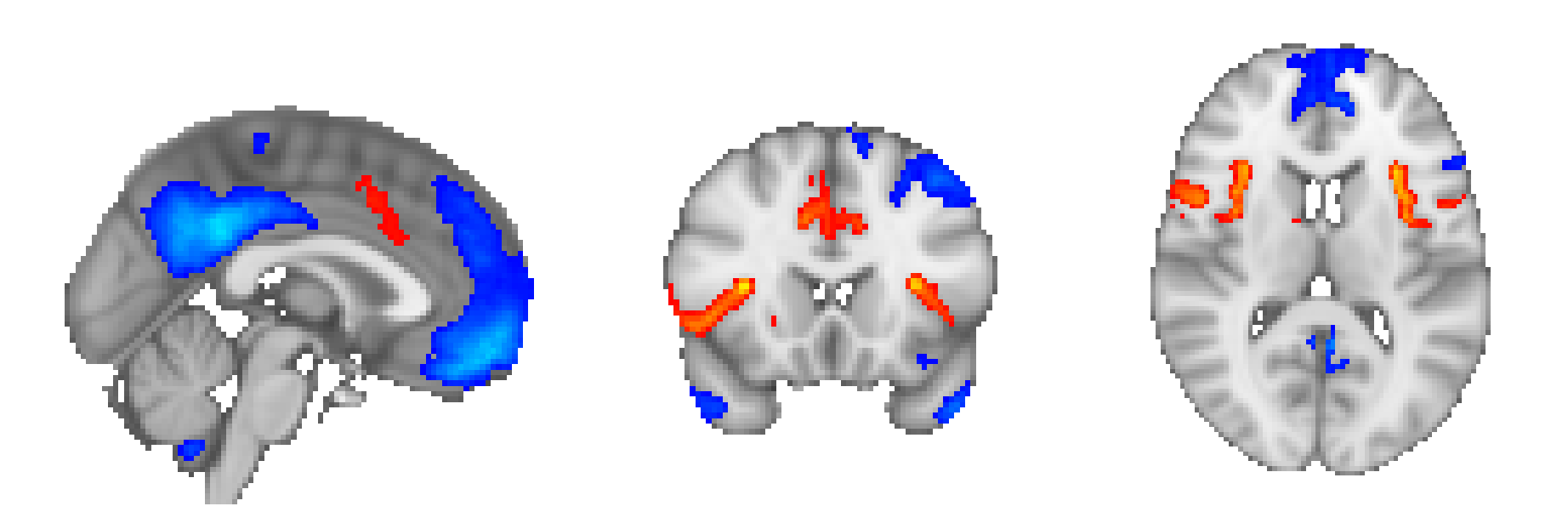}}}
      %{\includegraphics[width=.87\linewidth]{figures/7_old.png}}
        \end{minipage}
\begin{minipage}{.49\linewidth}
\centering
 \subcaptionbox{Component $4: $$\rho = -0.07$, $-\ln(p) = 8.77$}
 {\makebox[0.87\linewidth][c]{\includegraphics[width=.65\linewidth]
    {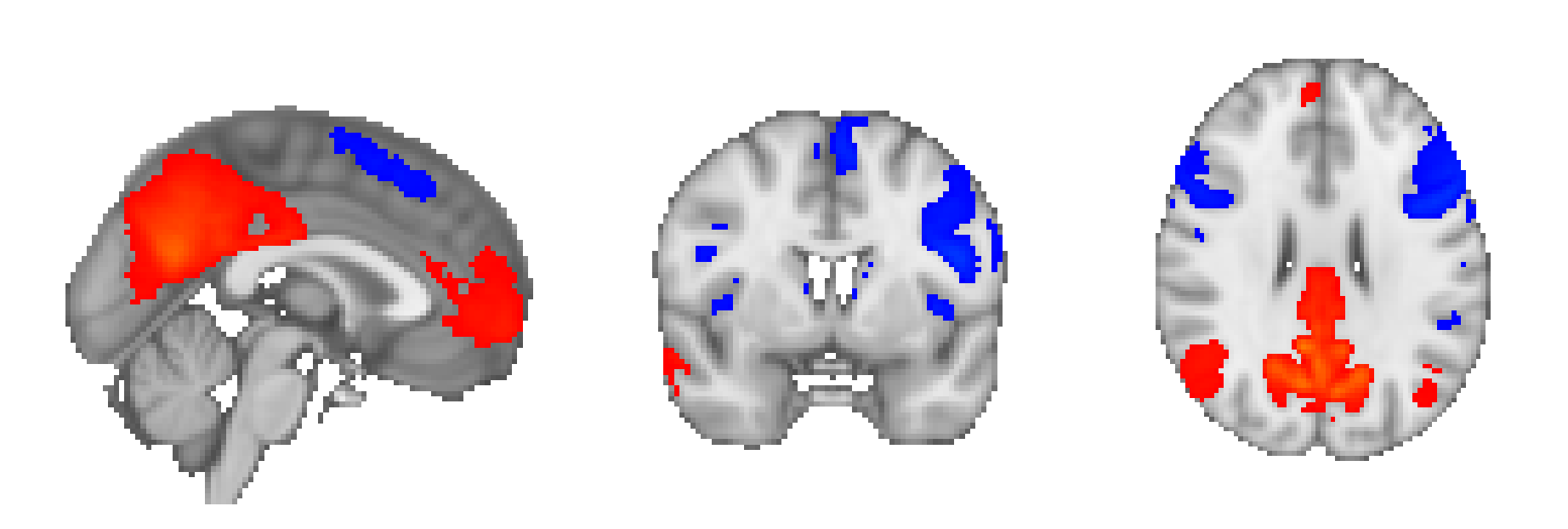}}}
      %{\includegraphics[width=.87\linewidth]{figures/7_new.png}}
  \end{minipage}
\caption{(a) Anticorrelation between Default Mode Network (DMN) and Salience Network (SN) and (b) Anticorrelation between Default Mode Network (DMN) and Central Executive Network (CEN). Red and blue regions are anti-correlated with each other but are correlated among themselves. The colors are not associated with negative or positive correlation since they can be swapped without affecting the final inference. \label{fig:SN} }
\end{figure}

\paragraph{Age prediction.} We used subject specific information ($\mathbf{\Lambda}$) having total $k_1 + k_2$ features from the two layers to predict age of each subject. We used Bootstrap-aggregated (bagged) decision trees to perform regression with $400$ trees for each site separately. Table \ref{tbl:mae_h2} shows the average and standard deviation of $10$ fold cross validation mean absolute error (MAE) across site for varied values of $k_1$ and $k_2 = 6$. We decided to perform age prediction of each site separately because the age is confounded by the site. The correlation between age and site is $0.24$ and reduction in site effects would reduce the prediction capability in the pooled setting.
\begin{table}[t!]
\caption{Mean absolute error ($k_2=6$)\label{tbl:mae_h2}}
\centering
        \begin{tabular}{ lcccc }
 %\hline
 %\multicolumn{5}{|c|}{Comparison by varying number of components}\\
 \toprule
 {Method} & {$k_1=10$} & {$k_1=15$}  & {$k_1=20$} & {$k_1=25$} \\
 \midrule
hSCP & $6.472\pm 1.417$   & $6.440\pm 1.470$   &  $6.418\pm 1.484$ &   $6.401\pm 1.478$\\
%Adv. hSCP & $9\pm0.028$ &  $9\pm0.021$  &  $9\pm0.023$ &    $9\pm0.019$ \\
%rshSCP & $9\pm0.024$ &   $9\pm0.021$ &    $9\pm0.014$ &   $9\pm0.018$ \\
Adv. rshSCP & $6.475\pm 1.250$ & $6.439\pm 1.411$ & $6.421\pm 1.454$ &  $6.403\pm 1.474 $ \\
 \bottomrule
\end{tabular}
\end{table}
\begin{table}[t!]
\caption{Split-sample reproducbility on real dataset ($k_2=6$). \label{tbl:real_repro_h2}}
\centering
        \begin{tabular}{ lcccc }
 %\hline
 %\multicolumn{5}{|c|}{Comparison by varying number of components}\\
 \toprule
 {Method} & {$k_1=10$} & {$k_1=15$}  & {$k_1=20$} & {$k_1=25$} \\
 \midrule
hSCP & $0.691\pm0.034$   & $0.688\pm0.034$   &  $0.677\pm0.029$ &   $0.668\pm0.032$\\
ComBat hSCP & $0.670\pm0.026$   & $0.664\pm0.028$   &  $0.635\pm0.030$ &   $0.626\pm0.028$\\
Adv. hSCP & $0.701\pm0.026$ &  $0.696\pm0.029$  &  $0.681\pm0.028$ &    $0.679\pm0.031$ \\
rshSCP & $0.776\pm0.027$ &   $0.748\pm0.029$ &    $0.722\pm0.032$ &   $0.721\pm0.024$ \\
Adv. rshSCP & $\mathbf{0.779\pm0.029}$ & $\mathbf{0.751\pm0.026}$ & $\mathbf{0.731\pm0.027}$ &  $\mathbf{0.732\pm0.025}$ \\
 \bottomrule
\end{tabular}
\end{table}
\begin{table}[t!]
\caption{Leave one site out reproducbility on real dataset($k_2=6$). \label{tbl:leave_repro_h2}}
\centering
        \begin{tabular}{ lcccc }
 %\hline
 %\multicolumn{5}{|c|}{Comparison by varying number of components}\\
 \toprule
 {Method} & {$k_1=10$} & {$k_1=15$}  & {$k_1=20$} & {$k_1=25$} \\
 \midrule
hSCP & $0.637\pm0.035$   & $0.600\pm0.036$   &  $0.578\pm0.031$ &   $0.560\pm0.034$\\
ComBat hSCP & $0.618\pm0.037$   & $0.589\pm0.033$   &  $0.543\pm0.035$ &   $0.521\pm0.032$\\
Adv. hSCP & $0.642\pm0.028$ &  $0.608\pm0.021$  &  $0.585\pm0.023$ &    $0.572\pm0.019$ \\
rshSCP & $0.701\pm0.032$ &   $0.691\pm0.034$ &    $0.668\pm0.031$ &   $0.659\pm0.029$ \\
Adv. rshSCP & $\mathbf{0.703\pm0.033}$ & $\mathbf{0.695\pm0.035}$ & $\mathbf{0.672\pm0.030}$ &  $\mathbf{0.666\pm0.031}$ \\
 \bottomrule
\end{tabular}
\end{table}
\paragraph{Association with brain aging.} We computed spearman correlation of age ($>60$) with $\mathbf{\Lambda}_1$ and $\mathbf{\Lambda}_2$. We then calculated p-values for the hypothesis test of no correlation against the alternative hypothesis of a nonzero correlation and are converted to $-\ln(p)$, where $\ln$ is log base $2$. Table \ref{tbl:corr_pval_k1} and \ref{tbl:corr_pval_k2} displays spearman correlation and  negative log $p$-value. The total number of subjects with age greater than $60$ is $2746$. In the case of negative log base 2, if the value is greater than $2.99$ then we consider it statistically significant, equivalent to $p$-value less than $0.05$. Figure \ref{fig:SN} shows anti-correlation between DMN, and Salience Network (SN) and Central Executive Network (CEN), and negative correlation with age. Previously, \citet{nagel2011load} has found a decrease in the functional coupling between DMN and the premotor cortex, corroborating our results.

\begin{table}[t!]
  \centering
     \caption{Spearman correlation ($\rho$) and p-value of age ($>60$) with $\mathbf{\Lambda}_1$ computed from hSCP (a) and Adv. rshSCP (b). 
    \label{tbl:corr_pval_k1}}
  \begin{tabular}{p{0.05cm}cccccccccccc}
    \toprule
    &     & {$1$}   & {$2$} & {$3$} & {$4$}  & {$5$}   & {$6$} & {$7$} & {$8$} & {$9$} & {$10$}  \\
    \midrule
    \multirow{ 2}{*}{a} & $\rho$&    $0.07$ &   $0.0$  &  $0.09$ & $-0.12$ & $-0.02$ & $-0.07$ & $-0.15$ & $-0.04$ & $-0.06$ &   $0.02$  
    \\
    &$-\ln(p)$&  $9.28$ &   $0.27$  &  $15.0$ & $35.3$ & $1.57$ & $9.32$ & $33.3$ & $3.10$ & $5.98$ &   $1.27$     \\ 
     \midrule
        \multirow{ 2}{*}{b} & $\rho$&    $0.05$ &   $-0.02$  &  $0.11$ & $-0.07$ & $-0.13$ & $-0.05$ & $-0.03$ & $0.0$ & $-0.02$ &   $0.02$ 
    \\
    &$-\ln(p)$&    $5.29$ &   $0.96$  &  $20.0$ & $8.77$ & $28.6$ & $5.27$ & $2.46$ & $0.16$ & $1.67$ &   $1.44$   \\
    \bottomrule
  \end{tabular}
\end{table}
\begin{table}[t!]
  \centering
      \caption{Spearman correlation ($\rho$) and p-value of age ($>60$) with $\mathbf{\Lambda}_2$ computed from hSCP and Adv. rshSCP. 
    \label{tbl:corr_pval_k2}}
  \begin{tabular}{ccccccc}
    \toprule
    &     & {\RNum{1}}   & {\RNum{2}} & {\RNum{3}} & {\RNum{4}}   \\
    \midrule
    \multirow{ 2}{*}{hSCP} & $\rho$&    $0.02$ &   $-0.04$  &  $-0.08$ & $-0.13$ 
    \\
    &$-\ln(p)$&  $1.69$ &   $3.68$  &  $9.83$ & $24.7$  \\
     \midrule
        \multirow{ 2}{*}{Adv. rshSCP} & $\rho$&    $0.05$ &   $-0.05$  &  $-0.06$ & $-0.09$ 
    \\
    &$-\ln(p)$&  $4.95$ &   $5.34$  &  $6.98$ & $13.8$   \\   
    \bottomrule
  \end{tabular}

\end{table}

\iffalse
\subsection{Human Brain Networks}
\label{sec:supp_HBN}
In this section, we give details about the usage of various functional networks in our work. Below are the networks and regions or parts of the regions comprising them- 
\begin{enumerate}[leftmargin=*]
\item Default Mode Network (DMN)- Medial prefrontal cortex, posterior cingulate cortex/precuneus, angular gyrus and lateral temporal lobes \cite{raichle2015brain}.
    %Medial prefrontal cortex, posterior cingulate cortex, superior and inferior frontal gyrus, lateral temporal lobes, inferior parietal lobule  and the medial temporal lobes \cite{raichle2015brain}.
\item Dorsal Attention Network (DAN) consists of Frontal eye fields, ventral frontal region, middle temporal, inferior parietal sulcus, superior parietal lobule and
dorsolateral prefrontal cortex \cite{fox2005human}. The network is active when a subject is at wakeful rest and not focusing on the outside acitivties. During wakeful rest, the subject could be daydreaming or mind-wandering. It is best known for being active when a person is not focused on the outside world and the brain is at wakeful rest, such as during daydreaming and mind-wandering. It can also be active during detailed thoughts related to external task performance
\item Salience Network (SN)- Anterior insula, dorsal anterior cingulate cortex,  amygdala, dorsomedial thalamus and hypothalamus \cite{seeley2007dissociable}.
\item  Central Executive Network (CEN)- Dorsolateral prefrontal cortex, posterior parietal cortex, intraparietal sulcus \cite{zanto2013fronto}.
\end{enumerate}
\fi
\section{Between-network connectivity in aging}
In this section, we discuss related work on changes in between-network connectivity in older adults. \citet{geerligs2014reduced} published one of the earliest studies on changes in between-network connectivity in older adults using seed-based analysis while participants performed an oddball task. They observed stronger connectivity (or weaker anticorrelations) between distinct functional networks. For example, they found age-related connectivity increases between the DMN, and the somatosensory and the CEN, which aligns with the results of current work. Several other studies reported similar results using different approaches \cite{ferreira2016aging,geerligs2015brain}. The DAN and DMN appear to show strong anticorrelations due to their presence in externally directed and internally directed cognition. \citet{spreng2016attenuated} used both resting and task data to show a decrease in anticorrelations between these networks in older compared to younger. 

The increase in connectivity between different networks can be thought of as a decrease in the segregation of networks. Previous studies have indicated that this decrease in segregation causes a reduction in the specialization of specialized networks, affecting information processing of the human brain \cite{liem2019functional}. \citet{grady2016age} analyzed the connections between DMN, DAN and CEN networks and observed a lower index of segregation in older as compared to young. Our results also indicate a decrease in anticorrelation between various networks, which can be thought of decrease in the segregation of networks, resulting in reorganization of the human brain in old age.

\end{document}